\documentclass[manuscript,screen]{acmart}
\AtBeginDocument{%
  }

\setcopyright{acmlicensed}
\copyrightyear{2025}
\acmYear{2025}
\acmDOI{XXXXXXX.XXXXXXX}
\acmISBN{978-1-4503-XXXX-X/2025/06}

\usepackage{url}
\usepackage{verbatim}
\usepackage{graphicx}
\usepackage{caption}
\usepackage[labelfont=bf]{caption} % 图编号加粗
\usepackage{multirow}
\usepackage{booktabs}
\usepackage{hyperref}
\usepackage{adjustbox}

\usepackage[ruled]{algorithm2e}

\usepackage{hyperref}
\usepackage{enumitem}

\begin{document}

%%
%% The "title" command has an optional parameter,
%% allowing the author to define a "short title" to be used in page headers.
\title{Robust Federated Learning against Noisy Clients via Masked Optimization}

%%
%% The "author" command and its associated commands are used to define
%% the authors and their affiliations.
%% Of note is the shared affiliation of the first two authors, and the
%% "authornote" and "authornotemark" commands
%% used to denote shared contribution to the research.
% \author{Ben Trovato}
% \authornote{Both authors contributed equally to this research.}
% \email{trovato@corporation.com}
% \orcid{1234-5678-9012}
% \author{G.K.M. Tobin}
% \authornotemark[1]
% \email{webmaster@marysville-ohio.com}
% \affiliation{%
%   \institution{Institute for Clarity in Documentation}
%   \city{Dublin}
%   \state{Ohio}
%   \country{USA}
% }

\author{Xuefeng Jiang}
\affiliation{%
  \institution{Institute of Computing Technology, Chinese Academy of Sciences \& University of Chinese Academy of Sciences}
  \city{Beijing}
  \country{China}}
\email{jiangxuefeng21b@ict.ac.cn}

\author{Tian Wen}
\affiliation{%
  \institution{Institute of Computing Technology, Chinese Academy of Sciences \& University of Chinese Academy of Sciences}
  \city{Beijing}
  \country{China}}
\email{wentian24s@ict.ac.cn}

\author{Zhiqin Yang}
\affiliation{%
  \institution{Department of Electronic Engineering, The Chinese University of Hong Kong}
  \city{Hong Kong}
  \country{China}}
\email{yangzqccc@link.cuhk.edu.hk}

\author{Lvhua Wu}
\affiliation{%
  \institution{Institute of Computing Technology, Chinese Academy of Sciences \& University of Chinese Academy of Sciences}
  \city{Beijing}
  \country{China}}
\email{wulvhua24s@ict.ac.cn}

\author{Yufeng Chen}
\affiliation{%
  \institution{Institute of Computing Technology, Chinese Academy of Sciences \& University of Chinese Academy of Sciences}
  \city{Beijing}
  \country{China}}
\email{chenyufeng17b@ict.ac.cn}

\author{Sheng Sun}
\affiliation{%
  \institution{Institute of Computing Technology, Chinese Academy of Sciences}
  \city{Beijing}
  \country{China}}
\email{sunsheng@ict.ac.cn}

\author{Yuwei Wang}
\affiliation{%
  \institution{Institute of Computing Technology, Chinese Academy of Sciences}
  \city{Beijing}
  \country{China}}
\email{ywwang@ict.ac.cn}

\author{Min Liu}
\affiliation{%
  \institution{Institute of Computing Technology, Chinese Academy of Sciences \& University of Chinese Academy of Sciences}
  \city{Beijing}
  \country{China}}
\email{liumin@ict.ac.cn}

%
% \footnote{Corresponding Author: Min Liu.}

\renewcommand{\shortauthors}{Xuefeng Jiang et. al.}

%%
%% By default, the full list of authors will be used in the page
%% headers. Often, this list is too long, and will overlap
%% other information printed in the page headers. This command allows
%% the author to define a more concise list
%% of authors' names for this purpose.

% \author{Anonymous Authors}
% \renewcommand{\shortauthors}{Anonymous Authors}

%%
%% The abstract is a short summary of the work to be presented in the
%% article.
\begin{abstract}
In recent years, federated learning (FL) has made significant advance in privacy-sensitive applications. 
However, it can be hard to ensure that FL participants provide well-annotated data for training. The corresponding annotations from different clients often contain complex label noise at varying levels.
This label noise issue has a substantial impact on the performance of the trained models, and clients with greater noise levels can be largely attributed for this degradation.
To this end, it is necessary to develop an effective optimization strategy to alleviate the adverse effects of these noisy clients.
In this study, we present a two-stage optimization framework, \textit{MaskedOptim}, to address this intricate label noise problem. 
The first stage is designed to facilitate the detection of noisy clients with higher label noise rates.
The second stage focuses on rectifying the labels of the noisy clients' data through an end-to-end label correction mechanism, aiming to mitigate the negative impacts caused by misinformation within datasets.
This is achieved by learning the potential ground-truth labels of the noisy clients' datasets via backpropagation.
To further enhance the training robustness, we apply the geometric median based model aggregation instead of the commonly-used vanilla averaged model aggregation. 
We implement sixteen related methods and conduct evaluations on three image datasets and one text dataset with diverse label noise patterns for a comprehensive comparison. 
Extensive experimental results indicate that our proposed framework shows its robustness in different scenarios. Additionally, our label correction framework effectively enhances the data quality of the detected noisy clients' local datasets. 
% Our codes will be open-sourced to facilitate related research communities.
Our codes are available via \href{https://github.com/Sprinter1999/MaskedOptim}{https://github.com/Sprinter1999/MaskedOptim}.
\end{abstract}

%%
%% The code below is generated by the tool at http://dl.acm.org/ccs.cfm.
%% Please copy and paste the code instead of the example below.
%%

% \ccsdesc[300]{Human-centered computing~Ubiquitous and mobile computing}

%%
%% Keywords. The author(s) should pick words that accurately describe
%% the work being presented. Separate the keywords with commas.

\received{23 May 2025}
% \received[revised]{12 March 2009}
% \received[accepted]{5 June 2009}

%%
%% This command processes the author and affiliation and title
%% information and builds the first part of the formatted document.
\maketitle

\section{Introduction}
\label{sec:intro}
In the era of the rapid advancement of Internet and mobile technologies, the proliferation of widely distributed mobile devices has led to an exponential surge in data generation. Leveraging this vast volume of data presents significant challenges for traditional centralized machine learning approaches, which typically necessitate the collection of data at the central server. Firstly, the data generated by edge devices often contains sensitive privacy information, and regulations such as GDPR \cite{gdpr} have been widely implemented to prohibit the cross-regional data sharing. Secondly, centralized methods entail the transfer of substantial data volumes, imposing considerable demands on data transmission infrastructure and storage resources.

To this end, federated learning (FL) emerges as new distributed machine learning method, aiming to leverage distributed data from mobile devices (named by clients hereafter) while protecting privacy \cite{fedavg,smartphone}. 
Specifically, it enables multiple clients to collaboratively train models without centralizing private data on the central server.
In recent years, FL achieves notable successes in real-world applications, particularly in highly sensitive fields such as healthcare \cite{bingjie,fedeye}. 
For example, medical institutions across different regions collaboratively train diagnostic models without sharing sensitive patient private data and health records.
Despite its significant potential to preserve privacy, FL still faces certain challenges in practice.
Distributed client-side data often exhibit non-independent and identically distributed (Non-IID) \cite{qinbinCrossSilo}  characteristics and numerous effective methods \cite{fedlf, fedcrac} are proposed to address this challenge in recent years. 
Nevertheless, another data related challenge has been long neglected in previous works. 
Lacking sufficient incentive mechanisms \cite{incentives}, data collected by clients usually contain inevitable label noise \cite{rhfl}, which evidently degrades the performance of trained models.
Specifically, clients can collect annotations by different low-cost methods, including using pretrained models \cite{fedlsr}, employing human annotators or crowd-sourcing \cite{robustfed} and web crawling with related context tags \cite{dora}. This results in data from different clients often containing varying levels of label noise.
% Therefore, clients in FL often have different label noise rates.

Therefore, developing robust training methods against label noise becomes a critical challenge in FL.
To address these challenges, some recent methods \cite{robustfl, fedlsr, fedrn, noro,feddshar,peijian} attempt to enhance the training robustness against label noise during training. 
Moreover, clients with higher noise levels tend to have a greater negative impact on local update process, as they introduce more misinformation into the trained model. 
Inspired by previous works \cite{noro}, we reckon it is necessary to exploit specific optimization mechanism to tackle these more unreliable clients.
In the meantime, few works explore the potential to correct the underlying noisy labels within client-side datasets which can further improve the motivation for these noisy clients to participate in FL \cite{fedelc}.

In this study, we focus on tackling this label noise issue in FL. Consistent with prior works \cite{noro,fedelc,dualoptim}, we categorize clients into two distinct groups: a detected clean group and a detected noisy client group. To mitigate the adverse effect of the detected noisy clients, we propose an end-to-end optimization framework, \textit{MaskedOptim}, to robustly handle the detected noisy clients with a tailored optimization mechanism, while allowing relatively clean clients to perform the vanilla local update process. Specifically, for detected noisy clients, we introduce a differentiable variable to track the label distribution estimation for each sample during local training.
Drawing inspiration from the small-loss technique \cite{coteaching}, which indicates that samples with smaller losses are more likely to be correctly labeled, we apply a valid mask to filter out samples with larger losses, thereby decreasing the noisy label supervision in the local update process. 
Additionally, to further minimize the negative impact of noisy clients, we adopt a robust aggregation strategy using the geometric median weights instead of traditional averaged weights. To sum up, our main contributions can be summarized as follows:

\begin{itemize}[leftmargin=0.5cm]
\item We introduce \textit{MaskedOptim}, a robust FL framework to tackle the clients with varying label noise rates. The proposed framework can simultaneously conduct robust local training and progressively refine the labels of client-side datasets which can effectively improve the data quality of participants.
\item Our proposed framework firstly distinguishes noisy clients from relatively clean ones with the assistance of the two-component Gaussian mixture model. For detected noisy clients, we design an end-to-end optimization strategy by introducing a learnable distribution variable of each sample. We exploit valid masking to decrease the negative supervision from possible original noisy labels. We further aggregate the models with the geometric median weights. 
\item Extensive experimental results on three image datasets and one text dataset of three types of label noise demonstrate the robustness of our proposed \textit{MaskedOptim} against sixteen baseline methods. Meanwhile, our framework empirically shows its superiority on the label correction. Our codes will be open-sourced to facilitate future research. 
\end{itemize}

\section{Related Works}
\label{sec:rw}
% Have checked by Xuefeng 25.03.28
Herein we briefly introduce some recent advance in related research aspects, including federated learning (FL), noisy label learning (NLL), and federated noisy label learning (FNLL).

\subsection{Federated Learning}

Federated learning (FL) enables distributed clients to collaboratively train a shared model without the exchange of local raw data. 
One major challenge lies in that distributed clients often possess non-independent and identically distributed (Non-IID) data. 
This characteristic causes substantial divergence in model updates among clients, thereby restricting model convergence and degrading overall performance \cite{fedtrip}. 
In recent years, many efforts has been proposed to tackle the problem of data heterogeneity \cite{fedcrac, fedlf, fedtrip, fedbiad,logitsfusion,wentian} and achieve relatively satisfactory performance. 
However, most existing works overlook another key challenge regarding data quality \cite{fnbench}. 
Due to the high overhead and limited incentives \cite{incentives} of annotating high-quality datasets, it is nearly impossible to ensure that all client data are cleanly labeled.
Consequently, datasets often contain diverse degrees of label noise.
Furthermore, FL prioritizes the privacy and thus prevents the central server from directly accessing client data, which makes inspecting the label quality of clients even harder \cite{jingjing}. 
Deep neural networks can memorize incorrect labels, which negatively impacts model convergence and generalization \cite{coteaching,coteaching+}, which has also been witnessed in the context of FL \cite{fnbench,fedlsr}. 
Therefore, tackling noisy labels effectively in FL remains a critical challenge to solve \cite{fnbench}.

\subsection{Noisy Label Learning}

To mitigate the impact of noisy labels, various centralized noisy label learning (NLL) methods  are proposed in recent years. 
This section discusses several representative NLL methods which can be divided into two research line, including sample selection approach and robust training approach.
Sample selection approach aims to reduce the impact of noisy labels by selecting high-reliability data. 
Co-teaching \cite{coteaching} uses two peer neural networks, each selecting a subset of samples during training to pass to the other network.
This approach assumes that low-loss samples are more likely to be correctly labeled, thereby mitigating the impact of noisy labels. 
For another research line, robust training approach improves noise resistance by designing noise-robust loss functions or adjusting training mechanisms. 
Symmetric cross entropy (SCE) \cite{symmetricce} combines cross entropy with reverse cross entropy, improving the model resilience to label noise. 
Ghosh et. al. proposes the mean absolute value (MAE) of error is inherently robust to label noise \cite{mae}.
Generalized Cross Entropy Loss (GCE) \cite{gce} introduces an adjustable parameter to reduce the impact of noisy labels, improving the generalization of deep neural networks. 
Finally, holistic robust training methods further enhance model robustness by integrating multiple techniques. 
DivideMix \cite{dividemix} combines collaborative training in peer networks \cite{coteaching}, MixUp data augmentation, and the semi-supervised learning framework MixMatch \cite{mixmatch}, demonstrating its adaptability to noisy data. 
One recent experimental study  discusses some common techniques to improve the training robustness \cite{unleashing}. 

Above centralized NLL methods consider less about some FL's inherent challenges or limitations like different label noise rates across clients, data heterogeneity, privacy constraints, and communication overhead.
Therefore, the challenge of noisy labels in FL still requires specific solutions to enhance robustness and performance \cite{fnbench}.

\subsection{Federated Noisy Label Learning}
In the context of Federated Learning (FL), the inherent discrepancies in label quality across clients invariably give rise to label noise, which significantly undermines the efficacy of model training. To combat this challenge, a series of pioneering efforts have been proposed in the literature. This section critically reviews several prominent federated noisy label learning (FNLL) approaches designed to handle label noise in FL environments.

Robust aggregation methods predate FNLL approaches as the initial attempts to mitigate the impact of low-quality or adversarial clients. These methods focus on strengthening the model aggregation process to ensure reliable global model updates. For instance, Median \cite{median} reduces the influence of extreme parameter values by computing the median of each parameter across client updates, thereby minimizing the detrimental effects of outliers. RFA \cite{rfa} extends this concept by employing the geometric median to aggregate model updates, effectively neutralizing the impact of anomalous parameter values.
TrimmedMean \cite{trimmedmean} ranks client parameter updates and removes a predefined percentage of the highest and lowest values before calculating the average, providing a simple yet effective mechanism for outlier suppression. Meanwhile, Krum \cite{krum} is specifically tailored to address byzantine faults. It selects the client update that is closest to the geometric center of the update set, thereby safeguarding the global model against malicious or faulty contributions from Byzantine participants. These techniques collectively form the cornerstone of robust FL frameworks, enabling reliable model training in the presence of noisy data and untrusted clients. 

Then let us discuss some representative FNLL methods. 
Robust FL\cite{robustfl} collects class centroids from clients to conduct class-wise semantic alignment during the local update procedure. 
The collected centroids from clients are aggregated at the server and then broadcast to the clients. Note that the centroids carry informative clues about the private data.
FedLSR \cite{fedlsr} introduces robust local  optimization techniques inspired by entropy minimization and self-supervised learning \cite{entropy} without extra sensitive information transmitting. 
FedRN \cite{fedrn} holds a server-side client model pool. 
Given a certain client, the server can retrieve reliable neighbor models which have similar data distributions or relatively clean data in the model pool to assist the local training process.
FedNoRo \cite{noro} initially divide clients into the clean group and the noisy group through a warm-up phase, then applying robust knowledge distillation between the global model and local models.
FLR \cite{flr} simultaneously exploits the global model and the local model to further assign pseudo labels during the local updating.
Based on FedNoRo, FedELC \cite{fedelc} designs specific fine-grained label correction framework for these detected noisy clients.
This work proposes an early intuition that there exists some possibility that we can robustly perform the training process and simultaneously improve the data quality of clients, which inspires our work. 
In this work, we identify the bottleneck of prior work FedELC, and propose more robust local optimization and exploit more effective robust model aggregation mechanism, which will be elaborated later in Section \ref{sec:method}. 
Another work FedNed \cite{fedned} considers more extreme scenario, where some malicious noisy clients (i.e., byzantine participants) hold high-level label noise rate. 
It utilizes the output of malicious clients' models as negative teachers to conduct inverse knowledge distillation, which can be studied in future works of communities.
Fed$A^3$I \cite{fedaaai} studies the label noise for the medical image segmentation task in FL.
Previous FNLL works mostly focus on the image related tasks, and FedDSHAR \cite{feddshar} is an early attempt to investigate the noisy label issue for time-series classification.
We also compare some previous  well-known related methods regarding the methodology design and experimental settings in Table \ref{tab:comparison_prev}.
More related knowledge can be found in previous surveys FNBench \cite{fnbench} and FedNoisy \cite{fednoisy}.

\begin{table}[tbp]
\caption{Comparison of related methods regarding  properties of methodology (\textbf{M}) and experimental settings (\textbf{E}) : (\textbf{M1}) Sample or client selection (\textbf{M2}) Robust aggregation, (\textbf{M3}) Explicit label correction, (\textbf{M4}) Requiring extra auxiliary dataset,  (\textbf{E1}) Data heterogeneity, (\textbf{E2}) Designed for label noise, (\textbf{E3}) Label noise of varying levels across clients, (\textbf{E4}) Real-world systematic label noise (see Section \ref{sec:labelnoise}) and (\textbf{E5}) Different Modalities.}
\label{tab:comparison_prev}
\begin{adjustbox}{width=0.8\columnwidth,center}
\begin{tabular}{c|c|cccc|ccccc}
\toprule \toprule
\textbf{Methods}  &\textbf{Venue}& \textbf{M1} & \textbf{M2} & \textbf{M3}  &\textbf{M4}& \textbf{E1} & \textbf{E2} & \textbf{E3} & \textbf{E4}  & \textbf{E5} \\ \midrule
FedAvg \cite{fedavg}  &AISTATS&  &  &   && $\checkmark$ &  & &  &\\
FedProx \cite{fedprox} &MLsys&  &  &   && $\checkmark$ &  & &  &\\
FedExP \cite{fedexp} &ICLR&  &  &   && $\checkmark$ &  &  &  &\\ \midrule
TrimmedMean \cite{trimmedmean}  &ICML&  & $\checkmark$ &   &&  & $\checkmark$ & &  &\\ 
Krum \cite{krum}  &NeurIPS&  & $\checkmark$ &   &&  & $\checkmark$ &  &  &\\
RFA \cite{rfa}  &IEEE TSP&  & $\checkmark$ &   &&  & $\checkmark$ &  &  &\\
Median \cite{median}  &ICML&  & $\checkmark$ &   &&  & $\checkmark$ & &  &\\ \midrule
Co-teaching \cite{coteaching}  &NeurIPS& $\checkmark$ &  &   &&  & $\checkmark$ & &  &\\
Co-teaching+ \cite{coteaching+} &ICML& $\checkmark$ &  &   &&  & $\checkmark$ & &  &\\
Joint Optim \cite{jointopt} &CVPR&  &  & $\checkmark$  &&  & $\checkmark$ &  & $\checkmark$ &\\
SELFIE \cite{selfie} &ICML& $\checkmark$ &  &   &&  & $\checkmark$ & &  $\checkmark$ &\\
Generalized Cross Entropy \cite{gce} &NeurIPS&  &  &   &&  & $\checkmark$ &  &  &\\
Mean Absolute Error \cite{mae} &AAAI&  &  &   &&  & $\checkmark$ &  &  &$\checkmark$ \\
Symmetric CE \cite{symmetricce} &ICCV&  &  &   &&  & $\checkmark$ &  &  &\\
DivideMix \cite{dividemix}  &ICLR& $\checkmark$ &  &   &&  & $\checkmark$ & &$\checkmark$  &\\ \midrule
Robust FL \cite{robustfl} & \small IEEE Intelligent Systems& $\checkmark$ &  &   &&  & $\checkmark$ & $\checkmark$ & $\checkmark$  &\\
RHFL \cite{rhfl} &CVPR&  &  &   &$\checkmark$ &  & $\checkmark$ &  &  &\\
FedLSR \cite{fedlsr} &ACM CIKM&  &  &   &&  & $\checkmark$ &  & $\checkmark$  &\\
FedRN \cite{fedrn} &ACM CIKM&  &  &   && $\checkmark$ & $\checkmark$ & $\checkmark$ &   &\\
FedNoRo \cite{noro} &IJCAI&  $\checkmark$ & $\checkmark$ &   && $\checkmark$ & $\checkmark$ & $\checkmark$ & &\\
FedNed \cite{fedned} &AAAI&  $\checkmark$ & &   &$\checkmark$ & $\checkmark$ & $\checkmark$ & $\checkmark$ & &\\
Fed$A^{3}$I \cite{fedaaai} &AAAI& $\checkmark$  & $\checkmark$ &   & & $\checkmark$ & $\checkmark$ & $\checkmark$ & &\\
FedELC \cite{fedelc}  &ACM CIKM& $\checkmark$ & $\checkmark$ & $\checkmark$  && $\checkmark$ & $\checkmark$ & $\checkmark$ & $\checkmark$  &\\ 
FLR \cite{flr}  & TMLR & $\checkmark$ & $\checkmark$ &   &  & $\checkmark$ & $\checkmark$ & $\checkmark$ &   &  \\ 
FedDSHAR \cite{feddshar}  &FGCS& $\checkmark$ & $\checkmark$ &   & $\checkmark$ & $\checkmark$ & $\checkmark$ & $\checkmark$ &   & $\checkmark$ \\ \midrule
Ours  & - & $\checkmark$ & $\checkmark$ & $\checkmark$  & & $\checkmark$ & $\checkmark$ & $\checkmark$ & $\checkmark$  & $\checkmark$ \\ \bottomrule
\end{tabular}
\end{adjustbox}
\end{table}

\section{Preliminary}

\subsection{Problem Definition}

Without loss of generality, we consider a distributed Federated Learning (FL) system comprising a central server and a set of  $N$ clients, denoted by $\mathcal{S}=\{1,2,\cdots , N\}$. Each client $k\in \mathcal{S}$ maintains a local dataset $\mathcal{D}_{k}=\left\{\left(x_{i}, \hat{y}_{i}\right)\right\}_{i=1}^{n_{k}}$ with $n_k$ samples. Note that the one-hot label $\hat{y}_{i}\in \{0,1 \}^{C}$ can be noisy, and we denote the unknown real ground-truth label is $y^*_i\in \{0,1 \}^{C}$, with $C$ being the number of classes. 

The overall objective of FL is to collaboratively train a global model parameterized by $\boldsymbol{w} \in \mathbb{R}^p$, where $p$ is the dimensionality of the model parameters, by optimizing the following global objective function:
\begin{equation}
\min_{\boldsymbol{w}} f(\boldsymbol{w}) := \sum_{k \in \mathcal{S}} \frac{n_k}{n} F_k(\boldsymbol{w}),
\end{equation}
where $n=\sum_{k\in \mathcal{S}}n_k$ is the total data quantity and $F_k(\boldsymbol{w})$ represents the local objective function for client $k$, defined as the expected loss over its local dataset $\mathcal{D}_k$:
\begin{equation}
F_k(\boldsymbol{w}) = \mathbb{E}_{(x, \hat{y}) \sim \mathcal{D}_k} \left[ \ell_k(\hat{y}, f_k(x; \boldsymbol{w_k})) \right].
\end{equation}
where $\ell_k$ is the loss function on the local dataset of $k$-th client and $f_{k}(x;w_k)\in \mathbb{R}^C$ is the local prediction on each sample $x$ with the local trained model parameterized by  $w_k$. Following the standard FL framework \cite{fedavg}, the training process proceeds over $T$ global communication rounds.

In each round $t \in \{0, 1, \dots, T-1\}$, the server firstly selects a group of clients $\mathcal{S}_t \subseteq \mathcal{S}$ to perform local update  for $E$ local epochs, and then it aggregates the updated local models from $\mathcal{S}_t$ to form the updated global model $w^{t+1}$:
\begin{equation}
\label{eq:fedavg}
    w^{t+1}=\sum_{k\in S_{t}}\frac{n_{k}}{n_t}w_{k}^{t},
\end{equation}
where $n_t$ indicates the total number of data of all selected clients $\mathcal{S}_t$  selected in the round $t$.

\subsection{Label Noise Modeling}
\label{sec:labelnoise}
Herein we discuss three types of label noise patterns in this work, following the definition of  previous studies \cite{fnbench,fedelc,dualoptim,fedrn}: 

\textbf{Synthetic label noise.}
The label transition matrix $\mathcal{M}$ is often employed to model the manually injected synthetic label noise in local datasets. 
Here, $\mathcal{M}_{i,j} = \text{flip}(\hat{y}=j|y^*=i)$ indicates that the ground-truth label $y^*$ is transformed from the clean class $i$ to the noisy class $\hat{y}=j$.
There exist two frequently-used structures for the matrix $\mathcal{M}$: symmetric flipping and asymmetric (or pairwise) flipping \cite{coteaching}.
Symmetric flipping implies that the original class label is transformed to any incorrect class label with an equal probability. 
In the case of pairwise flipping, it means that the original class label is only transformed to another incorrect category.
In FL, given the maximum noise rate $\epsilon$, for each client $k$, the corresponding noise level increases linearly from $0$ to $\epsilon$ as the client index $k$ increases.
In addition to the symmetric and asymmetric label noise scenarios, in line with \cite{fedrn,fedelc, fnbench}, we  generate a mixed label noise scenario.
In this scenario, clients are partitioned into two equal groups. One half of the clients adhere to the symmetric label noise model, while the remaining half follow the asymmetric label noise model. 

\textbf{Human annotation errors caused label noise.}
In the prior research\cite{}, evaluations on datasets that mirror real-world human annotation errors have been underutilized. Recently, Amazon Mechanical Turk introduced the CIFAR-N datasets \cite{cifarn}, which gather labels exclusively from human annotators.
These datasets have been  employed in several recent studies \cite{chaos,fedelc,cifarn}.
Concerning specific iterations of the CIFAR-N datasets, this study utilizes the CIFAR-10-N-Worst version due to the relatively high noise rate (40.21\%). 

\textbf{Systematic label noise.} 
In practice, it is expensive to obtain high-quality labels for large-scale datasets.
One commonly-utilized method is to collect images from websites and yield annotations by filtering the surrounding context (e.g. image captions \cite{clip}) in the web page \cite{dora}. 
Clothing1M \cite{clothing1m} is a large-scale real-world dataset of 14 categories, which contains 1 million images of clothing with noisy labels crawled from several online shopping websites. The overall noise rate is about 39.46\% and contains unstructured complicated label noise.

\section{Method}
\label{sec:method}
\begin{figure}[htbp]
    \centering
    \includegraphics[width=\linewidth]{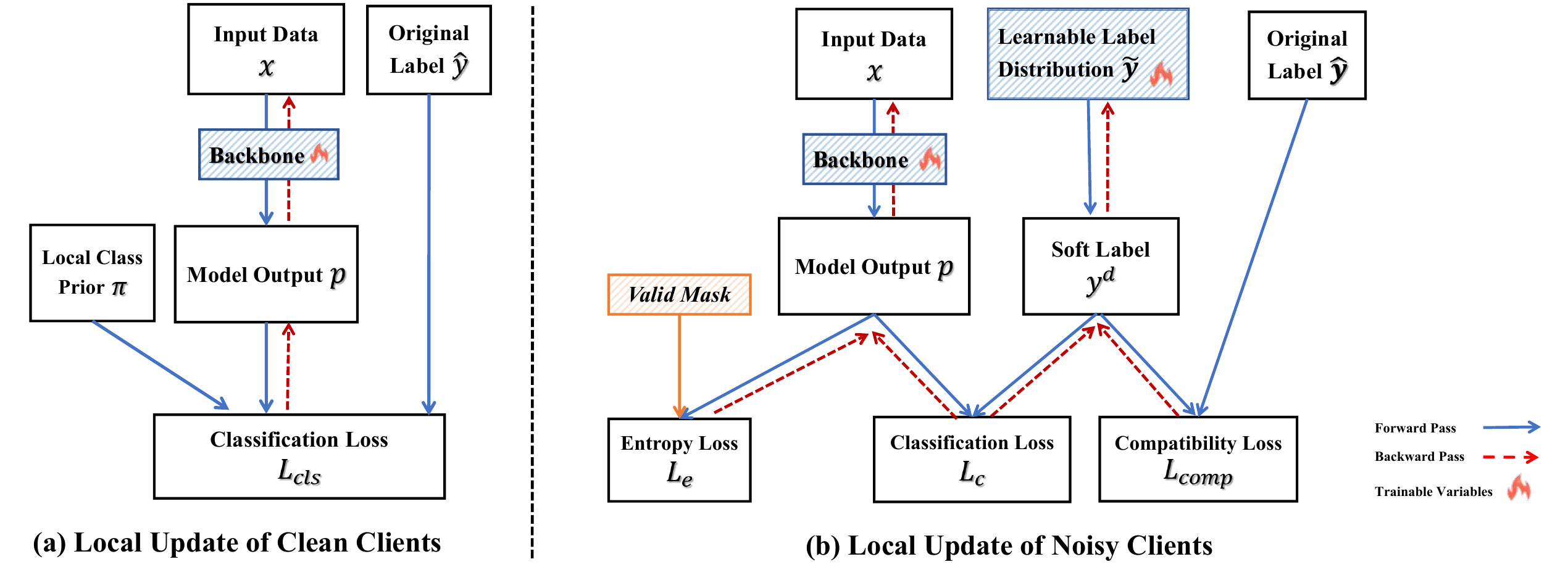}
    \caption{Illustration of \textit{MaskedOptim}.}
    \label{fig:framework}
\end{figure}

To collaborate clients with unknown and varying levels of label noise, we propose our \textit{MaskedOptim} framework as illustrated in Figure \ref{fig:framework}. During the first stage, all clients are orchestrated to follow the vanilla training procedure to warm up the global model. After warm-up epochs, we divide all clients into one relatively clean group and the comparatively noisy group with the assistance of the two-component Gaussian mixture models \cite{gmm}. 
For those detected noisy clients, we implement an improved masked end-to-end optimization mechanism to simultaneously update the local model and progressively update the estimation of possible ground-truth label for each sample.
During the whole training process, the server aggregate the local models from selected clients via the robust aggregation technique proposed in RFA \cite{rfa}. 

\subsection{Noisy Client Detection}
Our method is divided into two stages where the first stage lasts for $T_w$ global communication rounds to warm up the global model. 
During this stage, a warm-up model is firstly trained based on FedAvg \cite{fedavg}.
When the warm-up training ends, we exploit a two-component Gaussian Mixture Model (GMM) to divide total clients into a relatively clean group and a relatively noisy group.

During each round $t\in T_w$, the selected client $k\in \mathcal{S}_t$ updates its local model with the vanilla cross-entropy loss. Different from the original FedAvg \cite{fedavg}, we explicitly address the local data heterogeneity by incorporating class prior information across clients. Specifically, each selected client first computes the local class distribution $\pi$ based on its local dataset. This class distribution serves as a prior and is utilized to conduct \textbf{logit adjustment} \cite{LA} before computing the cross-entropy loss,
Prior studies \cite{noro,fedelc} have demonstrated the effectiveness of logit adjustment in addressing class imbalance and bias by encouraging the model to treat all classes more uniformly. Given a training sample $(x,\hat{y})$, the model yields the prediction output (called as logits) $p=f_k(x;w^t_k)$.
The logits are then adjusted using the prior as $p + log(\pi)$, and the cross-entropy loss $\mathcal{L}_{cls}$ is computed as

% \vspace{2pt}
\begin{equation}
\label{eq:cls}
    \mathcal{L}_{cls} = CE(p,\pi,\hat{y})
\end{equation}

After $T_w$ global training rounds, we perform the \textbf{ noisy client detection} procedure, which adopts a fine-grained and class-aware method for identifying potentially noisy client, following the approach in \cite{noro,fedelc}. Specifically, each client $k$ uses the global model $w^{T_w}$, obtained at the end of  warm-up stage, to compute the average loss for each class on its local dataset.  The average loss values of all classes $\ell_k = {(\ell_k^1, \ell_k^2,\ldots, \ell_k^C)} \in \mathbb{R}^C $ contain nothing about private information, and they will be transmitted to the server for further noisy client detection. By collecting the loss vectors from all clients, we construct a loss matrix $L=[\ell_1,\ell_2,\ell_3,\ldots,\ell_N] \in \mathbb{R}^{N\times C}$, which encapsulates per-class loss profiles across clients. Then, a two-component GMM is deployed onto $L$ to partition all $N$ clients into two subsets: the relatively clean group $\mathcal{S}_{clean}$ and the detected noisy group $\mathcal{S}_{noisy}$. More details can be seen available in \cite{noro}. 

\subsection{End-to-end Masked Optimization} 
\label{sec:stage2}
\begin{algorithm}[tbp] 
    \caption{Local Update Procedure}
    \label{algo}
    \LinesNumbered
    \KwIn{client $k$ , current global round $t$, global model $w_t$ }
    \KwOut{local trained model $w_t^k$}%输出
    $w_t^k \gets w_t$  \; 
    \If{t < $T_w$ or $k$ is clean client}
    {
     \textcolor{blue}{/* Vanilla training for clean clients or the warm-up stage*/}\\
     Compute local class distribution prior $\pi$ \; 
    \For{each local epoch $i$ from 1 to $E$}{
        \For{each batch $(x,\hat{y})$ }{
        % $o_1, o_2 = f(x;w_t^k), f(Augment(x);w_t^k)$\;
        % $o_2 = f(Augment(x);w_t^k)$\;
        % $p_1, p_2 = SoftMax(o_1), SoftMax(o_2)$ \; 
        % $\lambda \sim Beta(1,1)$ \;
        % $p = \lambda*p_1 + (1-\lambda)*p_2$  \;
        % $p_{s} = Sharpen(p,T)$ \;
        % \textcolor{blue}{/* Phase 2: FedAvg with FLR loss*/}\\
        $p = f(x; w_t^k)$           \; 
        $\mathcal{L}_{cls}$ = CE($p$, $\hat{y}$, $\pi$) ; \textcolor{blue}{// Eq. \ref{eq:cls}}  \\
        % $Loss_{reg} = SelfDistillation(o_1,o_2, T_d)$ \;
        % $Loss = Loss_{cls} + \gamma * Loss_{reg} $ \;
        Update $w_t^k$ with $\mathcal{L}_{cls}$ via \textit{Back Propagation} \;
        }
    }
    }
    \Else{  
    \textcolor{blue}{/* End-to-end label correction for noisy clients*/}\\
    Initialize $y_d$ with original label $\hat{y}$ \;
    \For{each local epoch $i$ from 1 to $E$}{
        \For{each batch B $(x,y)$ }{
        $p = f(x;w_t^k)$\;
        mask = valid\_mask(p,$\hat{y}$, filter rate $\tau$)  ; \textcolor{blue}{// Eq. \ref{eq:validmask}}  \\ 
        % \textcolor{blue}{/* Phase 2: FedAvg with FLR loss*/}\\
        $\mathcal{L}_{c} = CE(p,y^d)$ ;  \textcolor{blue}{// Eq. \ref{eq:classification}}  \\ 
        $\mathcal{L}_{comp} = Compatibility(\hat{y},\tilde{y})$  ; \textcolor{blue}{// Eq. \ref{eq:comp}}  \\
        $\mathcal{L}_{e} = Entropy(p)*mask$ ; \textcolor{blue}{// Eq. \ref{eq:entropy} and Eq. \ref{eq:entropy_v2}} \\ 
        $\mathcal{L} = \mathcal{L}_{c} + \alpha * \mathcal{L}_{comp} + \beta * \mathcal{L}_{e} $ ; \textcolor{blue}{// Eq. \ref{eq:all_loss}}  \\
        Update $w_t^k$ \& $\tilde{y}$ with $\mathcal{L}$ via \textit{Back Propagation} \;
        % Update $$ with $Loss$ via \textit{Back Propagation} \;
        }
    }
        \textcolor{blue}{/*Merge learned labels and updated prediction*/} \\
    \For{each local sample x}{
    $P_{model} = SoftMax(f(x; w_t^k))$  \;
    % $P_{predict} = SoftMax(P_{logit}) $\;
    $\tilde{y} = \frac{P_{model} + SoftMax(\tilde{y})}{2} \times K$ ;
    % \textcolor{blue}{// $y^{d}$  is also updated}
    \\
    $y_{estimate} = SoftMax(\tilde{y})$ ; \textcolor{blue}{// Estimate possible $y^{*}$}  \\
    }
    Pre-merge $w_t^k$ with the received $w_t$ ; \textcolor{blue}{// Mitigate weight drifts from the received global model in Eq. \ref{eq:merge}.} \\
    }
\end{algorithm}

We adopt different local update strategies for the two groups of clients $\mathcal{S}_{clean}$ and $\mathcal{S}_{noisy}$. For clients in $\mathcal{S}_{clean}$, local updates are conducted using the standard optimization objective defined in Eq. \ref{eq:cls}. Conversely, in order to more comprehensively utilize data from detected noisy clients $S_{noisy}$, the following optimization procedure is implemented . Drawing inspiration from \cite{pencil,fedelc}, we perform both robust local model updates and local data refinement through end-to-end correction of local labels.

We firstly introduce the concepts of hard/soft label for M-way classification. We denote $\mathbf{1}$ is a vector of all-ones. For the label of a certain sample, the one-hot hard label space is $\mathcal{H}=\{y:y\in\{0,1\}^C,\mathbf{1}^\top y=1\}$, and the soft label space is $\mathcal{S}=\{y:y\in[0,1]^m,\mathbf{1}^\top y=1\}$, which reflects a probabilistic label distribution. 
For a sample $(x,\hat{y})$ in the local dataset, we suppose it has the unknown ground-truth label $y^*$, and both $y^*$ and $\hat{y}$ is a one-hot hard label. We initialize a distribution $y^d \in \mathcal{S} = \{y:y\in[0,1]^C,1^\top y=1\}$ to denote our estimation of the underlying pseudo-ground-truth soft label distribution, which is initialized by the original $\hat{y}$ and can be gradually updated via back propagation:

\begin{equation}
    \tilde{y}=K\hat{y},
\end{equation}

\begin{equation}
    y^d=\operatorname{SoftMax}(\tilde{y}),
\end{equation}

We begin by introducing the concepts of hard and soft labels in the context of \textit{C}-way classification. Let $\mathbf{1}$ denote an all-ones vector. The hard label space is defined as $\mathcal{H}=\{y:y\in\{0,1\}^C,\mathbf{1}^\top y=1\}$, representing the conventional one-hot encoding where each sample belongs to exactly one class. In contrast, the soft label\textbf{ }space is defined as $\mathcal{S}=\{y:y\in[0,1]^C,\mathbf{1}^\top y=1\}$, which corresponds to a probabilistic distribution over the \textit{C} classes, allowing fractional class memberships that sum to one. Given a sample $(x,\hat{y})$ from the local dataset, we assume the existence of an unknown ground-truth label $y^*$, where both $y^*$and the provided label $\hat{y}$ are elements of the hard label space $\mathcal{H}$. To model an estimated soft label distribution, we introduce a distribution $y^d \in \mathcal{S}$ that serves as our \textit{pseudo-ground-truth} distribution. This distribution is initialized from the observed label  $\hat{y}$ and can be gradually updated during training via backpropagation as:

\begin{equation}
    \tilde{y}=K\hat{y},
\end{equation}
\begin{equation}
    y^d=\operatorname{SoftMax}(\tilde{y}),
\end{equation}
\
where $\tilde{y}$ is a differentiable variable subject to optimization and $K$ is a large constant.

% ($K = 10$ by directly following the original implementation of \cite{pencil}).

For noisy clients, the local optimization objective is composed of three terms. 
The first loss term is the basic classification loss $\mathcal{L}_{c}$. Instead of employing the original label $\hat{y}$, we calculate the classification loss between the model prediction $p$ and the learnable distribution $\tilde{y}$: 

\begin{equation}
\label{eq:classification}
    \mathcal{L}_{c} = CE(p,\tilde{y})
\end{equation}

The second loss is the compatibility regularization loss. In most scenarios wherein the noise levels of the majority of clients are not overly high, the original label $\hat{y}$ can still offer valuable supervision \cite{fedelc,pencil}. In other words, for the entire set of samples, the optimized distribution $\tilde{y}$ should not deviate substantially from the original labels $\hat{y}$. The compatibility regularization loss can be defined as: 

\begin{equation}
\label{eq:comp}
    \mathcal{L}_{comp} = Compatibility(\hat{y},\tilde{y}) = -\sum_{m=1}^C\hat{y}_{c}\log (y_{c}^d).
\end{equation}

The third term is the entropy regularization loss, which is widely used in NLL or FNLL methods \cite{fedlsr,robustfl,dividemix}, and semi-supervised learning methods \cite{mixmatch,entropy}. The main target is to encourage the model to output sharper and more confident prediction, which can be formulated as:
\begin{equation}
\label{eq:entropy}
     \mathcal{L}_e= Entropy(p) =-\sum_{c=1}^Cp^{c}\log(p^{c}),
\end{equation}
% \vspace{-1pt}
where $p^{c}$ is the model’s output softmax probability for the $c$-th class.

While the aforementioned loss components are formulated at the sample level, local model training is conducted on the batch level in practice. Since the original labels $\hat{y}$ can contain unreliable supervision,  we can not blindly require the model to yield sharper prediction for all samples in a training batch, even if they have high possibility to be incorrectly labeled.
Inspired by the small-loss trick \cite{coteaching,robustfl} which indicates the samples with smaller supervised loss are more likely to be correctly labeled. For each training batch $\mathcal{B}$, we keep the $\tau\%$ samples which have smaller cross entropy loss (sample-level $\mathcal{L}_c$ in essence) and attain the batch-level mask by
\begin{equation}
\label{eq:validmask}
mask = valid\_mask(\tau,\mathcal{B}). 
\end{equation}

Therefore, we modify the entropy loss to:
\begin{equation}
\label{eq:entropy_v2}
     \mathcal{L}_e = \mathcal{L}_e \cdot mask.
\end{equation}

Combining all the three terms together, we form triplet supervision for the detected noisy clients as follows:
\begin{equation}
    \label{eq:all_loss}
    \mathcal{L} = \mathcal{L}_{c} + \alpha * \mathcal{L}_{comp} + \beta * \mathcal{L}_{e},
\end{equation}
where $\alpha$ and $\beta$ represent the trade-off coefficients to balance these loss terms. $\mathcal{L}_{c}$ and $\mathcal{L}_{comp}$ provide supervisions to optimize the learnable $\tilde{y}$, thus, the optimization process of $\tilde{y}$ can be formulated as
\begin{equation}
\label{eq:backward_y}
    \tilde{y} = \tilde{y} - \eta*\nabla\tilde{y}
\end{equation}

Note that the the learning rate $\eta$ for $\tilde{y}$ differs from the rate associated with the optimized model $\theta$. A more detailed discussion regarding its selection will be presented in Section \ref{sec:ablation}. After $E$ epochs of local updating, an estimation of the corrected label can be obtained. This corrected label serves as an approximation of the potential ground-truth label $y^*$, and is derived through the fusion of two estimation methods: the prediction of the trained model and the updated $\tilde{y}$.
Since the optimization objective of a noisy client $k$ differs  from that of  clean clients, it may lead to model weight drift as discussed in \cite{fedtrip,moon}. To mitigate this issue, we reckon that the uploaded model weights of the noisy client $k$ should retain part of the global knowledge. This is achieved by pre-merging the local updated model $w_t^k$ and the received global model $w_t$:
\begin{equation}
    \label{eq:merge}
    w_t^k = \zeta w_t^k + (1-\zeta)w_t,
\end{equation}
where $\zeta$ denotes the pre-merging percentage of the trained model, and we provide its sensitivity study in Section \ref{sec:ablation}. 

\begin{algorithm}[htbp]
\caption{Robust aggregation with geometric median values}\label{algo1}
\LinesNumbered
\KwIn{models $w_i \in S_t$ for current communication round $t$, convergence threshold $\epsilon$, max iterations $max\_iter$}
\KwOut{Aggregated model $w_{global}$}
\For{each model parameter $k$}{
    Stack parameters from all clients: $W_k \leftarrow [w_1[k], w_2[k], ..., w_{|S_t|}[k]]$\;
    Initialize median: $\mu_k \leftarrow \text{mean}(W_k)$\;
    \For{$iter = 1$ to $max\_iter$}{
        Compute distances: $d_i \leftarrow \|W_k[i] - \mu_k\|_2$ for all $i$\;
        Calculate weights: $w_i \leftarrow \frac{1}{\max(d_i, \epsilon)}$\;
        Normalize weights: $w_i \leftarrow \frac{w_i}{\sum_j w_j}$\;
        Update median: $\mu_k^{new} \leftarrow \sum_i w_i \cdot W_k[i]$\;
        \If{$\|\mu_k^{new} - \mu_k\|_2 < \epsilon$}{
            break\;
        }
        $\mu_k \leftarrow \mu_k^{new}$\;
    }
    $w_{global}[k] \leftarrow \mu_k$\;
}
\Return $w_{global}$
\end{algorithm}

\subsection{Robust Aggregation via Geometric Median}
To further decrease the negative impacts caused by noisy clients and consider 
the model weight drifts issues that caused by the different optimization behaviors between clean and noisy clients, we robustly aggregate the updated models using the geometric median weights instead of traditional weighted average weights.
Inspired by \cite{rfa}, our key motivation is to automatically reduce the influence of outlier clients without requiring explicit identification, thereby enhancing robustness against malicious clients or those with poorly labled data. The pseudocodes are shown in Algorithm \ref{algo1}. Referring to previous works \cite{rfa,fnbench}, we assign \textit{$max\_iter$} to 10 and $\epsilon$ to $1e^{-5}$ in our implementation.

\section{Experiment}

\subsection{Experimental Setup}
\textbf{Datasets.} 
In this study, we investigate three types of label noise ( as detailed in Section \ref{sec:labelnoise}) and conduct experiments on both image and text dataset.
An overview of dataset statistics is summarized in Table \ref{tab:dataset}.
We apply standard normalization using the mean and standard deviation specific to each image dataset. To simulate data heterogeneity across clients,
we employ a Dirichlet distribution with concentration parameter $\gamma$ to partition the training data. A lower $\gamma$  value corresponds to a higher degree of data Non-IID.  For the image classification task, we set $\gamma$ to $0.5$ to introduce a moderate level of data heterogeneity following \cite{fedelc,fedrn}. 
For the text classification task on the AGNews dataset, we set $\gamma$ to 1.0 following \cite{fnbench}. For the CIFAR-10-N dataset \cite{cifarn}, we additionally conduct experiments with IID partitioning, since it is less explored in this FNLL research.

\begin{table}[htb]
  \centering
    \caption{Datasets and base models utilized in experiments. $\dagger$ denotes the pre-trained ResNet \cite{kaiming} on ImageNet \cite{imagenet} is used.}
  \label{tab:dataset}
  % \begin{adjustbox}{width=\columnwidth,center}
\begin{tabular}{c|c|c|c|c}
\toprule \toprule
\textbf{Dataset}                      & CIFAR-10 \cite{cifar} & CIFAR-10-N \cite{cifarn}        & AGNews \cite{agnews}               & Clothing1M \cite{clothing1m}       \\ \midrule
\textbf{Size of $\mathcal{D}$}        & 50,000   & 50,000  & 120,000                 & 1,000,000        \\
\textbf{Size of $\mathcal{D}_{test}$} & 10,000   & 10,000 & 7,600              & 10,000           \\
\textbf{Modality}                   & Image & Image & Text           & Image        \\ 
\textbf{Label Noise Type}                   & Synthetic & Human Annotation Errors & Synthetic           & Systematic        \\ 
\small \textbf{Total Communication Round $T$} & 120  &120 &60 &40 \\
\textbf{Base Model}                   & ResNet-18 \cite{kaiming} & ResNet-18 \cite{kaiming} & FastText \cite{fasttext}          & ResNet-50 \cite{kaiming} $\dagger$        \\ \bottomrule 
\end{tabular}
  % \end{adjustbox}

\end{table}

\textbf{Baseline methods.} We evaluate the state-of-the-art methods, categorized as follows:
\begin{itemize}[leftmargin=0.5cm]
    \item General federated learning (FL): FedAvg \cite{fedavg}, FedProx \cite{fedprox}, FedExP \cite{fedexp}.
    \item Robust aggregation: TrimmedMean \cite{trimmedmean}, Median \cite{median}, Krum \cite{krum}.
    \item Noisy label learning (NLL): Co-teaching \cite{coteaching}, Co-teaching+ \cite{coteaching+}, Joint Optim\cite{jointopt},  DivideMix \cite{dividemix}.
    \item Federated noisy label learning (FNLL): RoubstFL \cite{robustfl}, FedRN 
 \cite{fedrn}, FedLSR \cite{fedlsr}, FedNoRo \cite{noro},  FedELC \cite{fedelc}.
\end{itemize}

Unless otherwise specified, most hyper-parameters of these methods are configured favorably in line with the original literature and one previous FNLL benchmark study \cite{fnbench}. For clarity on key hyperparameter values, we provide detailed settings below: (1) General FL: The $\mu$ of FedProx \cite{fedprox} is fixed to 0.1 and $\epsilon$ of FedExP \cite{fedexp} is fixed to $1e^{-3}$; (2) Robust aggregation: Krum \cite{krum} and TrimmedMean \cite{trimmedmean} require to set the upper bound of the ratio of compromised or bad clients (denoted as $\kappa$). We select $\kappa$ to $0.3$ following \cite{fedelc,fnbench}. (3) NLL: The forget rate is fixed to 0.2 in Co-teaching\cite{coteaching} and Co-teaching+ \cite{coteaching+}. (4) FNLL: The warm-up epochs for RobustFL \cite{robustfed}, FedLSR \cite{fedlsr}, FedRN \cite{fedrn} are set to 20\% of total global training rounds. The $\gamma_e$ and $\gamma$ of FedLSR \cite{fedlsr} are set to $0.3$ and $0.4$. The reliable neighbor number is set as 1 for FedRN \cite{fedrn}. The forget rate of Robust FL is set to $0.2$.

\textbf{Implementation details.}
All main experiments are performed with Pytorch framework \cite{pytorch} on Nvidia$^{\circledR}$ RTX 3090 GPUs. For evaluation, class-wise averaged precision rate (@Pre) and recall rate (@Rec) are adopted as the main metrics, with the support of the scikit-learn package \cite{sklearn}. 
For a fair comparison, all experiments are averaged over 3 different random seeds.
Mixed precision training \cite{mixedPrecision} is employed to accelerate the model training process. 
In total, $N = 100$ clients are generated. 
In each $t$-th global communication round of FL, 10 clients (i.e., $|S_t|=10$) participate, which represents a standard experimental setup in FL, as described in \cite{fedavg,fedlsr,fedelc}. 
The local iteration epoch $E$ for each client is set to 5, and the local batch size is 64.
The SGD optimizer is chosen with a fixed learning rate of $0.01$, a momentum of $0.9$, and a weight decay of $5\times10^{-4}$, following \cite{fedelc}. 

\textbf{Hyper-parameter selection.}
Four hyper-parameters are involved in the our framework. These are the trade-off coefficients $\alpha$ and $\beta$ in Equation \ref{eq:all_loss}, the warm-up epochs $T_w$ of stage\#1, and the learning rate $\eta$ for updating the possible label distribution $\tilde{y}$. 
For the trade-off coefficients, $\alpha$ is set to $0.5$ and $\beta$ is set to $0.1$. 
The warm-up epoch $T_w$ is chosen as 30. 
The $\tau$ in Eq.\ref{eq:validmask} is mildly fixed to 80, referring to \cite{fnbench}.
Regarding the learning rate $\eta$ for updating label distribution estimation, it is fixed at 1000. 
We fix $\zeta$ in Eq.\ref{eq:merge} as 0.8 in our implementation.

\subsection{Analysis on Synthetic Noisy Datasets}
\label{sec:main_exp}
For the synthetic label noise scenario, we manually inject label noise to CIFAR-10 dataset for the image recognition task, and AGNews for the text classification task.
We list the experimental results of synthetic noisy datasets with manually injected noisy labels in Table \ref{tab:results_c10} and Table \ref{tab:agnews}.
For CIFAR-10, we find the proposed \textit{MaskedOptim} achieves more superiority in robustness for all scenarios, even if the label noise rate for each client is ranged from $0.0$ to $0.8$. We also find when the label noise rate is not very high ($0.0$ to $0.4$), most general FL methods (like FedAvg) and robust aggregation method (like median and RFA) show surprising robustness compared with other more complicated methods. This also supports the reason that we do not apply extra intricate optimization mechanisms on the detected clean clients.

\begin{table*}[htbp]
\caption{Precision (@Pre.) and recall rates (@Rec.) on synthetic noisy dataset CIFAR-10 with manually-injected noisy labels. Sym./Asym. refer to symmetric/asymmetric label noise. The \textbf{bold} denotes the best result and the \underline{underlined} denotes the best result in the corresponding  group.}
\label{tab:results_c10}
\resizebox{0.8\linewidth}{!}{
\begin{tabular}{c|cc|cc|cc|cc}
\toprule\toprule
\multirow{2}{*}{\textbf{Method}} & \multicolumn{2}{c|}{Sym. (0.0-0.4)} & \multicolumn{2}{c|}{Sym. (0.0-0.8)} & \multicolumn{2}{c|}{Asym. (0.0-0.4)} & \multicolumn{2}{c}{Mixed (0.0-0.4)} \\ \cmidrule{2-9} 
 & @Pre. & @Rec. & @Pre. & @Rec. & @Pre. & @Rec. & @Pre. & @Rec. \\ \midrule
FedAvg \cite{fedavg} & 72.10& 62.96& 49.44& 45.34& 74.26& 64.65& 73.66& 67.48\\
FedProx \cite{fedprox} & 69.51& \underline{63.42}& 47.96& 43.90& \underline{74.49}& \underline{64.68}& 73.85& 67.84\\
FedExp \cite{fedexp} & \underline{72.20} & 63.00& \underline{50.22}& \underline{45.52}& 74.28& 64.27& \underline{73.87}& \underline{67.93}\\ \midrule
TrimmedMean \cite{trimmedmean} & \underline{68.92} &60.95& 49.92& 42.83& 69.40& 57.12& 71.88& 59.72 \\
Krum \cite{krum} & 33.46 &36.27& 25.18& 28.44& 18.58& 24.79& 25.14& 28.39 \\
Median \cite{median} & 68.40& \underline{63.67}& \underline{52.35}& \underline{48.28}& \underline{72.09}& \underline{64.89}& \underline{72.01}&\underline{65.82}\\ \midrule
Co-teaching \cite{coteaching} &72.13 &\underline{66.42}& 48.20& 43.90& \underline{74.91}& \underline{64.45}&74.12 &\underline{68.12}\\
Co-teaching+ \cite{coteaching+} &58.07 &46.97& 57.05& 44.79 &46.73& 41.30& 55.71& 44.67 \\
Joint Optim \cite{jointopt} &57.71 &57.32 &52.40 &\underline{52.12}&55.20 &54.31 &57.80 &57.10 \\
SELFIE \cite{selfie} &\underline{72.63}& 64.49 &\underline{60.00}&50.21& 73.36& 60.12& \underline{74.20}& 64.10\\
DivideMix \cite{dividemix} &57.38 & 50.39 & 55.09 & 50.22 & 63.01 & 61.91& 62.27 & 58.38 \\ \midrule
Robust FL \cite{robustfl} &54.33 & 45.26 & 55.75 & 24.47 &54.18 & 42.96 & 53.98 & 43.48 \\
FedLSR \cite{fedlsr} &64.22 & 57.57& 62.39& 55.99& 66.56 & 55.88 & 60.14 & 53.21 \\
FedRN \cite{fedrn} & 58.92 &  46.56 & 54.67 & 50.24 & 55.14 & 41.29  & 55.78 & 47.83 \\
FedNoRo \cite{noro} &71.01 & 70.81 & 59.10 & 57.38 & 73.43 & 72.57 & 73.21 & 72.15 \\
FedELC \cite{fedelc} & 73.03 & 71.21  &60.21 &59.77& 74.68 &73.45 &74.56 &72.82  \\ 
Ours (\textit{MaskedOptim}) & \textbf{\underline{73.44}} & \textbf{\underline{71.31}} & \textbf{\underline{64.80}}& \textbf{\underline{61.66}}& \textbf{\underline{76.28}}& \textbf{\underline{73.53}}& \textbf{\underline{75.86}}& \textbf{\underline{72.91}}\\ \bottomrule
\end{tabular}
}
\end{table*}

Few previous works investigate the label noise issue in different modalities other than the image modality. To this end, we propose an early exploration for the text classification task with synthetic noisy labels.
For experiments on the AGNews dataset, we find our proposed \textit{MaskedOptim} still shows relatively satisfying robustness compared with other FNLL methods, while the general FL method FedProx \cite{fedprox} shows the overall best performance against different label noise patterns. We do not re-implement FedLSR, FedRN and DivideMix since these methods are mainly proposed to deal with noisy labels of image data and lack detailed adaptation on the natural language related task. To improve robustness when learning on different modalities is one of our future works.

\begin{table}[htbp]
\small
\caption{Evaluated F1-score (\%) on the  AGNews dataset for text classification.}
\centering
\small
% \resizebox{0.9\linewidth}{!}{
\begin{tabular}{ccccc}
\toprule \toprule
\multirow{2}{*}{\textbf{Method}} & \multicolumn{4}{c}{\textbf{Label Noise Patterns}} \\
 & \begin{tabular}[c]{@{}c@{}}Symmetric\\ 0.0-0.4\end{tabular} & \begin{tabular}[c]{@{}c@{}}Symmetric\\ 0.0-0.8\end{tabular} & \begin{tabular}[c]{@{}c@{}}Pairflip\\ 0.0-0.4\end{tabular} & \begin{tabular}[c]{@{}c@{}}Mixed\\ 0.0-0.4\end{tabular} \\ \midrule
FedAvg \cite{fedavg} & 63.13& 58.77& 59.59&59.52\\
FedProx \cite{fedprox} & \textbf{\underline{65.10}}& \textbf{\underline{61.82}}& \underline{61.15}&\textbf{\underline{63.66}}\\
FedExP \cite{fedexp} & 62.73& 56.07 & 60.79 & 61.05\\ \midrule
TrimmedMean \cite{trimmedmean} & 58.71& 51.72& 54.21&55.87\\
Krum \cite{krum} & 45.65 & 42.61& 44.37 &46.76\\
Median \cite{median} & \underline{62.96} & \underline{60.72}& \underline{61.06}&\underline{63.27}\\ \midrule
Co-teaching \cite{coteaching} &\underline{63.42} &\underline{58.67} &\textbf{\underline{62.77}} &\underline{62.11}\\
Co-teaching+ \cite{coteaching+} &61.07 &55.20 &58.45 &58.05\\
JointOptim \cite{jointopt} & 37.91 & 31.85 &36.34 & 38.48\\
SELFIE \cite{selfie} &37.85 &16.13 &27.35 &35.61\\ 
DivideMix \cite{dividemix} & -& -& -&-\\  \midrule
RFL \cite{robustfl} &30.72 &28.07 &33.25 &29.95 \\
FedLSR \cite{fedlsr} & -& -& -&-\\
FedRN \cite{fedrn} & - & - & - &-\\
FedNoRo \cite{noro} & 62.98& \underline{57.77}& 57.71& 60.86\\ 
FedELC \cite{fedelc} & 61.13 & 57.29& 57.51 & 60.12 \\ 
Ours (\textit{MaskedOptim})  & \underline{63.01} & 57.33  & \underline{59.51} & \underline{61.12} \\ \bottomrule
\end{tabular}
% }
\label{tab:agnews}
\end{table}

\subsection{Analysis on Datasets with  Human Annotation Errors}
\label{sec:humanoise}
In previous studies, most noisy labels are generated by manually injecting synthetic label noise. In this work, we consider incorporating noisy labels caused by human annotation errors. We use the CIFAR-10-N dataset \cite{cifarn}. The experimental results are shown in Table \ref{tab:cifar10-n}. From these results, we find our proposed method shows generally robustness than its counterparts. Meanwhile, when we compare with the  results in Table \ref{tab:results_c10}, we can find this type label noise can be easier to deal with than the synthetic label noise. We find the general FL methods also shows robustness over noisy labels caused by human annotation errors. This also indicates future works can consider more simple yet effective training mechanisms. 

\begin{table}[htbp]
\caption{Experimental results (\%) on CIFAR-10-N dataset. }
\label{tab:cifar10-n}
\small
% \begin{adjustbox}{width=\columnwidth,center}
\begin{tabular}{c|cc|cccc}
\toprule \toprule
\multirow{3}{*}{\textbf{Method}} & \multicolumn{2}{c|}{IID} & \multicolumn{4}{c}{Non-IID} \\ %\cline{2-7} 
 & \multicolumn{2}{c|}{-} & \multicolumn{2}{c|}{Dir. ($\gamma=1.0$)} & \multicolumn{2}{c}{Dir. ($\gamma=0.5$)} \\
 & @Pre. & @Rec. & @Pre. & \multicolumn{1}{c|}{@Rec.} & @Pre. & @Rec. \\ \midrule
FedAvg \cite{fedavg}& \underline{87.42}& 87.25 & 83.85 & \multicolumn{1}{c|}{82.79} & \underline{81.37}& 73.40 \\
FedProx \cite{fedprox} & 86.79 & 86.77 & \underline{83.97}& \multicolumn{1}{c|}{83.50} & 81.31 & \underline{77.09}\\
FedExP \cite{fedexp}& 87.36 & \underline{87.31}& 83.88 & \multicolumn{1}{c|}{\underline{83.71}} & 81.22 & 73.70 \\ \midrule
TrimmedMean \cite{trimmedmean}& 86.02 & 85.88 & 81.05 & \multicolumn{1}{c|}{78.02} & 75.75 & 65.72 \\
Krum \cite{krum}& 74.58 & 72.56 & 68.61 & \multicolumn{1}{c|}{61.29} & 44.16 & 36.84 \\
Median \cite{median} & \underline{86.10}& \underline{86.03}& \underline{83.03}& \multicolumn{1}{c|}{\underline{81.47}} & \underline{78.47}& \underline{71.82}\\ \midrule
Co-teaching \cite{coteaching}& \underline{84.61}& \underline{84.55}& \underline{79.76}& \multicolumn{1}{c|}{\underline{77.05}} & \underline{73.12}& \underline{61.59}\\
Co-teaching+ \cite{coteaching+}& 78.68 & 78.37 & 71.08 & \multicolumn{1}{c|}{65.12} & 60.12 & 50.78 \\
Joint Optim \cite{jointopt}& 78.22 & 78.35 & 68.09 & \multicolumn{1}{c|}{68.03} & 59.97 & 59.72 \\
SELFIE \cite{selfie}& 83.79 & 83.72 & 78.73 & \multicolumn{1}{c|}{74.52} & 73.66& 58.34 \\
DivideMix \cite{dividemix}& 76.77 & 75.22 & 66.29 & \multicolumn{1}{c|}{66.02} & 59.32 & 56.07 \\ \midrule
Robust FL \cite{robustfl}& 75.77 & 75.58 & 65.33 & \multicolumn{1}{c|}{63.71} & 55.35 & 51.88 \\
FedLSR \cite{fedlsr}& 82.24 & 81.95 & 77.35 & \multicolumn{1}{c|}{74.29} & 60.78 & 53.82 \\
FedRN \cite{fedrn}& 75.61 & 75.10 & 76.45 & \multicolumn{1}{c|}{70.87} & 47.30 & 46.20 \\
FedNoRo \cite{noro}& 87.24 & 87.20 & 83.90 & \multicolumn{1}{c|}{83.81} & \textbf{\underline{82.67}}& 81.29 \\ 
FedELC \cite{fedelc} & 87.67 & 87.47 & 84.29  & \multicolumn{1}{c|}{84.19} & 82.36 & 81.25 \\  
Ours (\textit{MaskedOptim}) & \textbf{\underline{87.79}} & \textbf{\underline{87.77}} & \textbf{\underline{84.79}}  & \multicolumn{1}{c|}{\textbf{\underline{84.49}}} & 81.51 & \textbf{\underline{81.35}} \\ \bottomrule
\end{tabular}
% \end{adjustbox}
\end{table}

\subsection{Analysis on the large-scale Clothing1M Dataset with systematic label noise}
For pre-processing on the large-scale real-world dataset, all images are resized to $224\times224$ and normalized. 
The global communication lasts for 40 rounds since we utilize the pre-trained ResNet-50 model \cite{kaiming}. 
Experimental results are shown in Table \ref{tab:testonclothing}. 
We additionally re-implement FedCorr \cite{fedcorr} and ELR \cite{elr} following the open-source codes.
For \#15, the learning rate is fixed to 0.1 to achieve higher performance by following \cite{fedlsr}. 
For Robust FL \cite{robustfl} and FedNoRo \cite{noro}, warm-up period lasts 10 global rounds. 
Experimental results verify the proposed \textit{MaskedOptim} achieves more reliable performance against its counterparts. However, we investigate related literature \cite{dividemix} and find the performance in the FL scenario is still lower than that in the centralized scenario, which indicate the robustness challenge caused by FL still requires large  future efforts.

\begin{table}[htbp]
\caption{Results on systematic noisy dataset Clothing1M with random partitioning.}
\label{tab:testonclothing}
\small
% \resizebox{\linewidth}{!}{
\centering
\begin{tabular}{cc}
\toprule \toprule
 \textbf{Method}                 & Best Test Accuracy (\%)\\ \midrule
 FedAvg \cite{fedavg}                & 69.26\\ 
 FedProx \cite{fedprox}                & 67.62\\ 
 FedExP \cite{fedexp}      & \underline{69.53} \\ \midrule
 TrimmedMean \cite{trimmedmean}                        & 68.57\\ 
 Krum \cite{krum}      & 67.32 \\ 
 Median \cite{median}                   & 69.00 \\ \midrule
 Co-teaching \cite{coteaching} & 69.77 \\ 
 Co-teaching+ \cite{coteaching+}      & 68.02 \\ 
 Joint Optim \cite{jointopt}      & \underline{70.31} \\ 
 SELFIE \cite{selfie}                  & 69.97\\  
 DivideMix \cite{dividemix}                        & 70.08\\ 
 ELR \cite{elr}      & 68.66 \\ \midrule
 RobustFL $\dagger$ \cite{robustfl}      & 71.77 \\ 
 FedLSR \cite{fedlsr}                        & 69.78\\ 
 FedRN \cite{fedrn}      & 68.60  \\ 
 FedCorr \cite{fedcorr}      & 69.92 \\
 FedNoRo \cite{noro}      & 70.62 \\
 FedELC \cite{fedelc}      & 71.64\\ 
 Ours (\textit{Masked Optim})       & \textbf{\underline{71.89}}\\
\bottomrule
\end{tabular}
% }
\end{table}

%Co-teaching has a hyperparameter $R(T)$ which cannot be priorly inferred in the practical federated learning system (discussed in section \ref{results}).

\subsection{Ablation \& Sensitivity Study}
\label{sec:ablation}

\textbf{Ablation on the masking mechanism and model aggregation.}
We conduct ablation experiments in Table \ref{tab:ablation_on_trick} by removing the masking mechanism in Eq. \ref{eq:entropy_v2} and using different model aggregation methods \cite{noro}. 
We find the overall performance gets degraded, which reflects the effectiveness of our incorporated techniques.

\begin{table}[htbp]
\caption{Results for Ablation study on the role of masking mechanism and different robust model aggregation techniques.}
\label{tab:ablation_on_trick}
% \begin{adjustbox}{width=\columnwidth,center}
\begin{tabular}{c|cccc}
\toprule\toprule
\multirow{2}{*}{Components}                                       & \multicolumn{4}{c}{Metric}        \\
                                                                  & @Pre. & @Rec. & @F1-score & @Acc. \\ \midrule
Ours & 73.43 &  71.31 & 72.33 & 72.41 \\ \midrule
Removing mask in Eq.\ref{eq:entropy_v2}                                 & 72.16 $\downarrow$    & 70.24 $\downarrow$     & 70.13 $\downarrow$         & 70.25 $\downarrow$    \\ \midrule
Model Averaging \cite{fedavg,fedlsr,fedrn}       & 72.62 $\downarrow$    & 71.75 $\uparrow$     & 71.46 $\downarrow$         & 71.74 $\downarrow$     \\
Median Aggregation \cite{median}                             & 69.84 $\downarrow$    & 65.44 $\downarrow$     & 65.75 $\downarrow$         & 65.43 $\downarrow$    \\
Distance-aware Aggregation \cite{noro,fedelc,dualoptim} & 72.86 $\downarrow$    & 70.09 $\downarrow$     & 69.48 $\downarrow$         & 70.10 $\downarrow$     \\ \bottomrule
\end{tabular}
% \end{adjustbox}
\end{table}

\begin{figure}[htbp]
    \centering
    \includegraphics[width=\linewidth]{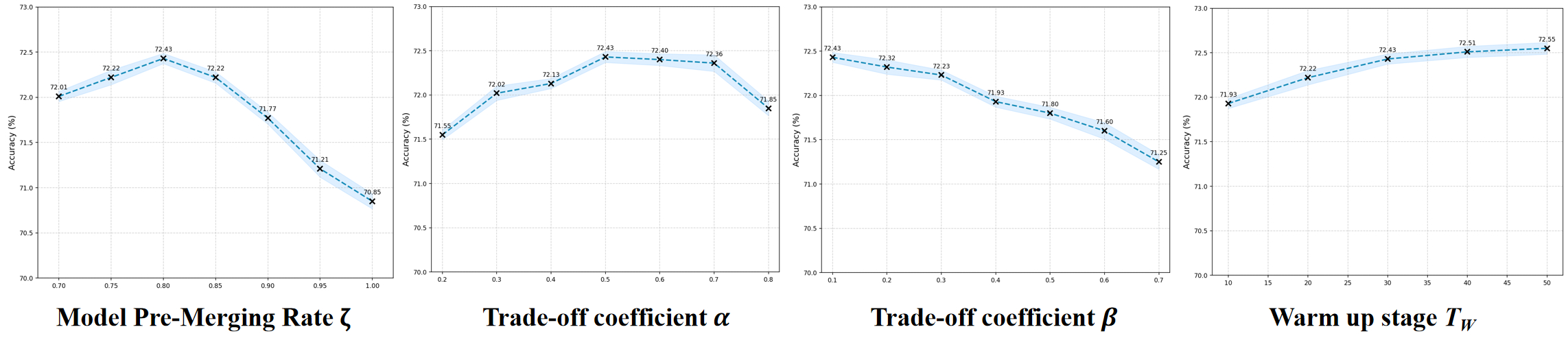}
    \caption{Visualization of the sensitivity  study.}
    \label{fig:sen}
\end{figure}

\textbf{Sensitivity of main hyper-parameters.} Herein we analyze the sensitivity of some main hyper-parameters included in this work. These hyper-paramters include the trade-off coefficients $\alpha$ and $\beta$ in Eq. \ref{eq:all_loss}, model pre-merging rate $\zeta$ and the warm-up epochs $T_w$. The results are visualized in Figure \ref{fig:sen}. We find the proposed \textit{MaskedOptim} shows its robustness over these four main hyper-parameters. We can also observe

\subsection{Analysis on Label Correction}
\label{sec:correction}
\begin{figure}[htbp]
    \centering
    \includegraphics[width=\linewidth]{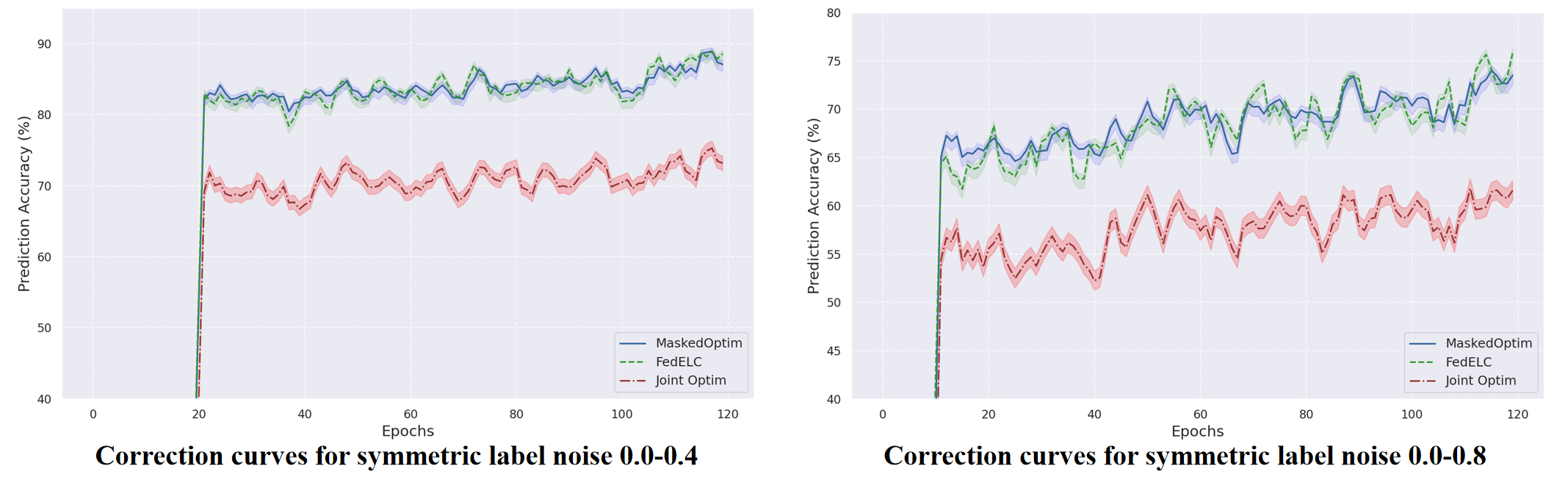}
    \caption{Visualization of the label correction performance.}
    \label{fig:corr}
\end{figure}

We reckon considering noisy labels in FL, we can not only improve the trained model's robustness against label noise but also try to conduct label correction. In Figure \ref{fig:corr}, we visualize the label correction performance of our proposed \textit{MaskedOptim}, FedELC \cite{fedelc} and Joint Optim \cite{jointopt} after $T_w$ global rounds for synthetic label noise scenarios.
We use our estimated label $y_{estimate}$ in Algorithm \ref{algo} and compare it with the true ground-truth label. 
We backup the true labels before adding label noise to the dataset to conveniently compute this correction accuracy. 
Results demonstrate that our method shows higher label estimation accuracy than JointOptim and achieves similar performance with FedELC.
Additionally, participants can take fewer efforts and time to compare the original annotated labels $\hat{y}$ with our estimated label $y_{estimate}$ and re-label some inconsistent samples ($\hat{y} \neq y_{estimate}$) to further improve the local data quality.

\section{Conclusion \& Discussion}
\label{sec:discuss}
In this work, we propose a robust federated learning framework \textit{MaskedOptim} to tackle the pervasive noisy labels within clients. 
Specifically, we divide the clients into the clean group and the noisy group after the first stage warm-up training.
After client division, we focus on the noisy clients and aim to progressively correct the labels of noisy clients via an end-to-end paradigm. 
Detailedly, a differentiable label distribution is introduced for each sample to guide the label correction process, which can provide another estimation of the possible ground-truth label. 
To decrease the negative impacts of the noisy labels, we construct a valid mask to filter out the possible misinformation from imperfect label supervision to further enhance the robustness of the trained model. 
To further mitigate the weight divergence in the model aggregation phase, we exploit the geometric median based model aggregation mechanism to decrease adverse impacts caused by deviated model parameters.

In essence, this work is largely inspired by previous work FedELC \cite{fedelc} which is an early attempt to correct the labels of noisy clients. Based on FedELC, we carefully revisit related literature and organize some well-known methods in Table \ref{tab:comparison_prev}, enhance the training robustness by designing more robust training mechanism with the masking technique on detected noisy clients and more robust model aggregation. 
Meanwhile, different from most FNLL works \cite{fedelc,fedlsr,noro,fedned}, we explore on the natural language processing task instead of solely evaluating only on the vision tasks. 
Experiments empirically show our improvement over the previous FedELC.
We encourage future studies can further explore the generalization on different model architectures \cite{vgg} and different learning tasks like dense prediction \cite{fedaaai,fedia}, semi-supervised setting \cite{zhiqin}, multi-modal learning \cite{bingjie} and sequence-level prediction \cite{feddshar}. 
In the meantime, some extreme scenarios which FedNed \cite{fedned} can be explored to tackle the extreme malicious FL participants.

% \subsection{Future Work}
For future works, we aim to explore some interesting yet less considered approaches to tackle the label noise issue.
Some more advanced and effective methods like reliable data generation \cite{fd,laf,yerui,transd} can be considered to help increase the robustness against the noisy data.  
We notice existing studies on two directions, semi-supervised learning and label noise learning, share some common correlations. 
The underlying intuition may lie in that pseudo labels in semi-supervised learning can be viewed as imperfect labels in noisy label learning and the noisy labels can provide more self-supervised information which are often utilized in semi-supervised learning. 
Perhaps existing methods in one direction can bring more inspirations for another direction, and we have witnessed some early attempts \cite{fedia,fedlsr,feddshar}.

\section*{Acknowledgements}
This work is supported by the National Natural Science Foundation of China (No. 62072410, No. 62072436), the National Key Research and Development Program of China (2021YFB2900102). 
We thank Yida Bai from Southwest University for feedback. 
We also appreciate the valuable feedback from the anonymous reviewers.
The corresponding author of this work is Min Liu.

\bibliographystyle{ACM-Reference-Format}
\bibliography{main}

%%% -*-BibTeX-*-
%%% Do NOT edit. File created by BibTeX with style
%%% ACM-Reference-Format-Journals [18-Jan-2012].

\begin{thebibliography}{72}

%%% ====================================================================
%%% NOTE TO THE USER: you can override these defaults by providing
%%% customized versions of any of these macros before the \bibliography
%%% command.  Each of them MUST provide its own final punctuation,
%%% except for \shownote{} and \showURL{}.  The latter two
%%% do not use final punctuation, in order to avoid confusing it with
%%% the Web address.
%%%
%%% To suppress output of a particular field, define its macro to expand
%%% to an empty string, or better, \unskip, like this:
%%%
%%% \newcommand{\showURL}[1]{\unskip}   % LaTeX syntax
%%%
%%% \def \showURL #1{\unskip}           % plain TeX syntax
%%%
%%% ====================================================================

\ifx \showCODEN    \undefined \def \showCODEN     #1{\unskip}     \fi
\ifx \showISBNx    \undefined \def \showISBNx     #1{\unskip}     \fi
\ifx \showISBNxiii \undefined \def \showISBNxiii  #1{\unskip}     \fi
\ifx \showISSN     \undefined \def \showISSN      #1{\unskip}     \fi
\ifx \showLCCN     \undefined \def \showLCCN      #1{\unskip}     \fi
\ifx \shownote     \undefined \def \shownote      #1{#1}          \fi
\ifx \showarticletitle \undefined \def \showarticletitle #1{#1}   \fi
\ifx \showURL      \undefined \def \showURL       {\relax}        \fi
% The following commands are used for tagged output and should be
% invisible to TeX
\providecommand\bibfield[2]{#2}
\providecommand\bibinfo[2]{#2}
\providecommand\natexlab[1]{#1}
\providecommand\showeprint[2][]{arXiv:#2}

\bibitem[Berthelot et~al\mbox{.}(2019)]%
        {mixmatch}
\bibfield{author}{\bibinfo{person}{David Berthelot}, \bibinfo{person}{Nicholas Carlini}, \bibinfo{person}{Ian~J. Goodfellow}, \bibinfo{person}{Nicolas Papernot}, \bibinfo{person}{Avital Oliver}, {and} \bibinfo{person}{Colin Raffel}.} \bibinfo{year}{2019}\natexlab{}.
\newblock \showarticletitle{MixMatch: {A} Holistic Approach to Semi-Supervised Learning}. In \bibinfo{booktitle}{\emph{Advances in Neural Information Processing Systems 32: Annual Conference on Neural Information Processing Systems 2019, NeurIPS 2019, December 8-14, 2019, Vancouver, BC, Canada}}, \bibfield{editor}{\bibinfo{person}{Hanna~M. Wallach}, \bibinfo{person}{Hugo Larochelle}, \bibinfo{person}{Alina Beygelzimer}, \bibinfo{person}{Florence d'Alch{\'{e}}{-}Buc}, \bibinfo{person}{Emily~B. Fox}, {and} \bibinfo{person}{Roman Garnett}} (Eds.). \bibinfo{pages}{5050--5060}.
\newblock
\urldef\tempurl%
\url{https://proceedings.neurips.cc/paper/2019/hash/1cd138d0499a68f4bb72bee04bbec2d7-Abstract.html}
\showURL{%
\tempurl}


\bibitem[Blanchard et~al\mbox{.}(2017)]%
        {krum}
\bibfield{author}{\bibinfo{person}{Peva Blanchard}, \bibinfo{person}{El~Mahdi~El Mhamdi}, \bibinfo{person}{Rachid Guerraoui}, {and} \bibinfo{person}{Julien Stainer}.} \bibinfo{year}{2017}\natexlab{}.
\newblock \showarticletitle{Machine Learning with Adversaries: Byzantine Tolerant Gradient Descent}. In \bibinfo{booktitle}{\emph{Advances in Neural Information Processing Systems 30: Annual Conference on Neural Information Processing Systems 2017, December 4-9, 2017, Long Beach, CA, {USA}}}, \bibfield{editor}{\bibinfo{person}{Isabelle Guyon}, \bibinfo{person}{Ulrike von Luxburg}, \bibinfo{person}{Samy Bengio}, \bibinfo{person}{Hanna~M. Wallach}, \bibinfo{person}{Rob Fergus}, \bibinfo{person}{S.~V.~N. Vishwanathan}, {and} \bibinfo{person}{Roman Garnett}} (Eds.). \bibinfo{pages}{119--129}.
\newblock
\urldef\tempurl%
\url{https://proceedings.neurips.cc/paper/2017/hash/f4b9ec30ad9f68f89b29639786cb62ef-Abstract.html}
\showURL{%
\tempurl}


\bibitem[Deng et~al\mbox{.}(2009)]%
        {imagenet}
\bibfield{author}{\bibinfo{person}{Jia Deng}, \bibinfo{person}{Wei Dong}, \bibinfo{person}{Richard Socher}, \bibinfo{person}{Li-Jia Li}, \bibinfo{person}{Kai Li}, {and} \bibinfo{person}{Li Fei-Fei}.} \bibinfo{year}{2009}\natexlab{}.
\newblock \showarticletitle{Imagenet: A large-scale hierarchical image database}. In \bibinfo{booktitle}{\emph{2009 IEEE conference on computer vision and pattern recognition}}. Ieee, \bibinfo{pages}{248--255}.
\newblock


\bibitem[{European Parliament} and {Council of the European Union}(2016)]%
        {gdpr}
\bibfield{author}{\bibinfo{person}{{European Parliament}} {and} \bibinfo{person}{{Council of the European Union}}.} \bibinfo{year}{2016}\natexlab{}.
\newblock \bibinfo{booktitle}{\emph{Regulation ({EU}) 2016/679 of the {European} {Parliament} and of the {Council}}}.
\newblock
\urldef\tempurl%
\url{https://data.europa.eu/eli/reg/2016/679/oj}
\showURL{%
\tempurl}


\bibitem[Fang and Ye(2022)]%
        {rhfl}
\bibfield{author}{\bibinfo{person}{Xiuwen Fang} {and} \bibinfo{person}{Mang Ye}.} \bibinfo{year}{2022}\natexlab{}.
\newblock \showarticletitle{Robust Federated Learning with Noisy and Heterogeneous Clients}. In \bibinfo{booktitle}{\emph{{IEEE/CVF} Conference on Computer Vision and Pattern Recognition, {CVPR} 2022, New Orleans, LA, USA, June 18-24, 2022}}. \bibinfo{publisher}{{IEEE}}, \bibinfo{pages}{10062--10071}.
\newblock
\href{https://doi.org/10.1109/CVPR52688.2022.00983}{doi:\nolinkurl{10.1109/CVPR52688.2022.00983}}


\bibitem[Ghosh et~al\mbox{.}(2017)]%
        {mae}
\bibfield{author}{\bibinfo{person}{Aritra Ghosh}, \bibinfo{person}{Himanshu Kumar}, {and} \bibinfo{person}{P~Shanti Sastry}.} \bibinfo{year}{2017}\natexlab{}.
\newblock \showarticletitle{Robust loss functions under label noise for deep neural networks}. In \bibinfo{booktitle}{\emph{Proceedings of the AAAI conference on artificial intelligence}}, Vol.~\bibinfo{volume}{31}.
\newblock


\bibitem[Grandvalet and Bengio(2004)]%
        {entropy}
\bibfield{author}{\bibinfo{person}{Yves Grandvalet} {and} \bibinfo{person}{Yoshua Bengio}.} \bibinfo{year}{2004}\natexlab{}.
\newblock \showarticletitle{Semi-supervised Learning by Entropy Minimization}. In \bibinfo{booktitle}{\emph{Advances in Neural Information Processing Systems 17 [Neural Information Processing Systems, {NIPS} 2004, December 13-18, 2004, Vancouver, British Columbia, Canada]}}. \bibinfo{pages}{529--536}.
\newblock
\urldef\tempurl%
\url{https://proceedings.neurips.cc/paper/2004/hash/96f2b50b5d3613adf9c27049b2a888c7-Abstract.html}
\showURL{%
\tempurl}


\bibitem[Han et~al\mbox{.}(2018)]%
        {coteaching}
\bibfield{author}{\bibinfo{person}{Bo Han}, \bibinfo{person}{Quanming Yao}, \bibinfo{person}{Xingrui Yu}, \bibinfo{person}{Gang Niu}, \bibinfo{person}{Miao Xu}, \bibinfo{person}{Weihua Hu}, \bibinfo{person}{Ivor~W. Tsang}, {and} \bibinfo{person}{Masashi Sugiyama}.} \bibinfo{year}{2018}\natexlab{}.
\newblock \showarticletitle{Co-teaching: Robust training of deep neural networks with extremely noisy labels}. In \bibinfo{booktitle}{\emph{Advances in Neural Information Processing Systems 31: Annual Conference on Neural Information Processing Systems 2018, NeurIPS 2018, December 3-8, 2018, Montr{\'{e}}al, Canada}}, \bibfield{editor}{\bibinfo{person}{Samy Bengio}, \bibinfo{person}{Hanna~M. Wallach}, \bibinfo{person}{Hugo Larochelle}, \bibinfo{person}{Kristen Grauman}, \bibinfo{person}{Nicol{\`{o}} Cesa{-}Bianchi}, {and} \bibinfo{person}{Roman Garnett}} (Eds.). \bibinfo{pages}{8536--8546}.
\newblock
\urldef\tempurl%
\url{https://proceedings.neurips.cc/paper/2018/hash/a19744e268754fb0148b017647355b7b-Abstract.html}
\showURL{%
\tempurl}


\bibitem[He et~al\mbox{.}(2016)]%
        {kaiming}
\bibfield{author}{\bibinfo{person}{Kaiming He}, \bibinfo{person}{Xiangyu Zhang}, \bibinfo{person}{Shaoqing Ren}, {and} \bibinfo{person}{Jian Sun}.} \bibinfo{year}{2016}\natexlab{}.
\newblock \showarticletitle{Deep Residual Learning for Image Recognition}. In \bibinfo{booktitle}{\emph{2016 {IEEE} Conference on Computer Vision and Pattern Recognition, {CVPR} 2016, Las Vegas, NV, USA, June 27-30, 2016}}. \bibinfo{publisher}{{IEEE} Computer Society}, \bibinfo{pages}{770--778}.
\newblock
\href{https://doi.org/10.1109/CVPR.2016.90}{doi:\nolinkurl{10.1109/CVPR.2016.90}}


\bibitem[Jhunjhunwala et~al\mbox{.}(2023)]%
        {fedexp}
\bibfield{author}{\bibinfo{person}{Divyansh Jhunjhunwala}, \bibinfo{person}{Shiqiang Wang}, {and} \bibinfo{person}{Gauri Joshi}.} \bibinfo{year}{2023}\natexlab{}.
\newblock \showarticletitle{FedExP: Speeding Up Federated Averaging via Extrapolation}. In \bibinfo{booktitle}{\emph{The Eleventh International Conference on Learning Representations, {ICLR} 2023, Kigali, Rwanda, May 1-5, 2023}}. \bibinfo{publisher}{OpenReview.net}.
\newblock
\urldef\tempurl%
\url{https://openreview.net/pdf?id=IPrzNbddXV}
\showURL{%
\tempurl}


\bibitem[Jiang et~al\mbox{.}(2024a)]%
        {fnbench}
\bibfield{author}{\bibinfo{person}{Xuefeng Jiang}, \bibinfo{person}{Jia Li}, \bibinfo{person}{Nannan Wu}, \bibinfo{person}{Zhiyuan Wu}, \bibinfo{person}{Xujing Li}, \bibinfo{person}{Sheng Sun}, \bibinfo{person}{Gang Xu}, \bibinfo{person}{Yuwei Wang}, \bibinfo{person}{Qi Li}, {and} \bibinfo{person}{Min Liu}.} \bibinfo{year}{2024}\natexlab{a}.
\newblock \showarticletitle{FNBench: Benchmarking Robust Federated Learning against Noisy Labels}.
\newblock \bibinfo{journal}{\emph{Authorea Preprints}} (\bibinfo{year}{2024}).
\newblock


\bibitem[Jiang et~al\mbox{.}(2025a)]%
        {dualoptim}
\bibfield{author}{\bibinfo{person}{Xuefeng Jiang}, \bibinfo{person}{Peng Li}, \bibinfo{person}{Sheng Sun}, \bibinfo{person}{Jia Li}, \bibinfo{person}{Lvhua Wu}, \bibinfo{person}{Yuwei Wang}, \bibinfo{person}{Xiuhua Lu}, \bibinfo{person}{Xu Ma}, {and} \bibinfo{person}{Min Liu}.} \bibinfo{year}{2025}\natexlab{a}.
\newblock \showarticletitle{Refining Distributed Noisy Clients: An End-to-end Dual Optimization Framework}.
\newblock \bibinfo{journal}{\emph{Authorea Preprints}} (\bibinfo{year}{2025}).
\newblock


\bibitem[Jiang et~al\mbox{.}(2025b)]%
        {transd}
\bibfield{author}{\bibinfo{person}{Xuefeng Jiang}, \bibinfo{person}{Yuan Ma}, \bibinfo{person}{Pengxiang Li}, \bibinfo{person}{Leimeng Xu}, \bibinfo{person}{Xin Wen}, \bibinfo{person}{Kun Zhan}, \bibinfo{person}{Zhongpu Xia}, \bibinfo{person}{Peng Jia}, \bibinfo{person}{XianPeng Lang}, {and} \bibinfo{person}{Sheng Sun}.} \bibinfo{year}{2025}\natexlab{b}.
\newblock \showarticletitle{TransDiffuser: End-to-end Trajectory Generation with Decorrelated Multi-modal Representation for Autonomous Driving}.
\newblock \bibinfo{journal}{\emph{arXiv preprint arXiv:2505.09315}} (\bibinfo{year}{2025}).
\newblock


\bibitem[Jiang et~al\mbox{.}(2024b)]%
        {fedelc}
\bibfield{author}{\bibinfo{person}{Xuefeng Jiang}, \bibinfo{person}{Sheng Sun}, \bibinfo{person}{Jia Li}, \bibinfo{person}{Jingjing Xue}, \bibinfo{person}{Runhan Li}, \bibinfo{person}{Zhiyuan Wu}, \bibinfo{person}{Gang Xu}, \bibinfo{person}{Yuwei Wang}, {and} \bibinfo{person}{Min Liu}.} \bibinfo{year}{2024}\natexlab{b}.
\newblock \showarticletitle{Tackling Noisy Clients in Federated Learning with End-to-end Label Correction}. In \bibinfo{booktitle}{\emph{Proceedings of the 33rd {ACM} International Conference on Information and Knowledge Management, {CIKM} 2024, Boise, ID, USA, October 21-25, 2024}}, \bibfield{editor}{\bibinfo{person}{Edoardo Serra} {and} \bibinfo{person}{Francesca Spezzano}} (Eds.). \bibinfo{publisher}{{ACM}}, \bibinfo{pages}{1015--1026}.
\newblock
\href{https://doi.org/10.1145/3627673.3679550}{doi:\nolinkurl{10.1145/3627673.3679550}}


\bibitem[Jiang et~al\mbox{.}(2022)]%
        {fedlsr}
\bibfield{author}{\bibinfo{person}{Xuefeng Jiang}, \bibinfo{person}{Sheng Sun}, \bibinfo{person}{Yuwei Wang}, {and} \bibinfo{person}{Min Liu}.} \bibinfo{year}{2022}\natexlab{}.
\newblock \showarticletitle{Towards Federated Learning against Noisy Labels via Local Self-Regularization}. In \bibinfo{booktitle}{\emph{Proceedings of the 31st {ACM} International Conference on Information {\&} Knowledge Management, Atlanta, GA, USA, October 17-21, 2022}}, \bibfield{editor}{\bibinfo{person}{Mohammad~Al Hasan} {and} \bibinfo{person}{Li~Xiong}} (Eds.). \bibinfo{publisher}{{ACM}}, \bibinfo{pages}{862--873}.
\newblock
\href{https://doi.org/10.1145/3511808.3557475}{doi:\nolinkurl{10.1145/3511808.3557475}}


\bibitem[Joulin et~al\mbox{.}(2017)]%
        {fasttext}
\bibfield{author}{\bibinfo{person}{Armand Joulin}, \bibinfo{person}{Edouard Grave}, \bibinfo{person}{Piotr Bojanowski}, {and} \bibinfo{person}{Tomas Mikolov}.} \bibinfo{year}{2017}\natexlab{}.
\newblock \showarticletitle{Bag of Tricks for Efficient Text Classification}. In \bibinfo{booktitle}{\emph{Proceedings of the 15th Conference of the European Chapter of the Association for Computational Linguistics: Volume 2, Short Papers}}.
\newblock
\href{https://doi.org/10.18653/v1/e17-2068}{doi:\nolinkurl{10.18653/v1/e17-2068}}


\bibitem[Kang et~al\mbox{.}(2023)]%
        {unleashing}
\bibfield{author}{\bibinfo{person}{Hui Kang}, \bibinfo{person}{Sheng Liu}, \bibinfo{person}{Huaxi Huang}, \bibinfo{person}{Jun Yu}, \bibinfo{person}{Bo Han}, \bibinfo{person}{Dadong Wang}, {and} \bibinfo{person}{Tongliang Liu}.} \bibinfo{year}{2023}\natexlab{}.
\newblock \showarticletitle{Unleashing the Potential of Regularization Strategies in Learning with Noisy Labels}.
\newblock \bibinfo{journal}{\emph{CoRR}}  \bibinfo{volume}{abs/2307.05025} (\bibinfo{year}{2023}).
\newblock
\href{https://doi.org/10.48550/arXiv.2307.05025}{doi:\nolinkurl{10.48550/arXiv.2307.05025}}
\showeprint[arXiv]{2307.05025}


\bibitem[Kim et~al\mbox{.}(2022)]%
        {fedrn}
\bibfield{author}{\bibinfo{person}{Sangmook Kim}, \bibinfo{person}{Wonyoung Shin}, \bibinfo{person}{Soohyuk Jang}, \bibinfo{person}{Hwanjun Song}, {and} \bibinfo{person}{Se{-}Young Yun}.} \bibinfo{year}{2022}\natexlab{}.
\newblock \showarticletitle{FedRN: Exploiting k-Reliable Neighbors Towards Robust Federated Learning}. In \bibinfo{booktitle}{\emph{Proceedings of the 31st {ACM} International Conference on Information {\&} Knowledge Management, Atlanta, GA, USA, October 17-21, 2022}}, \bibfield{editor}{\bibinfo{person}{Mohammad~Al Hasan} {and} \bibinfo{person}{Li~Xiong}} (Eds.). \bibinfo{publisher}{{ACM}}, \bibinfo{pages}{972--981}.
\newblock
\href{https://doi.org/10.1145/3511808.3557322}{doi:\nolinkurl{10.1145/3511808.3557322}}


\bibitem[Kim et~al\mbox{.}(2024)]%
        {flr}
\bibfield{author}{\bibinfo{person}{Taehyeon Kim}, \bibinfo{person}{Donggyu Kim}, {and} \bibinfo{person}{Se-Young Yun}.} \bibinfo{year}{2024}\natexlab{}.
\newblock \bibinfo{booktitle}{\emph{FLR: Label-Mixture Regularization for Federated Learning with Noisy Labels}}.
\newblock
\urldef\tempurl%
\url{https://openreview.net/pdf?id=Z8A3HDgS0E}
\showURL{%
\tempurl}


\bibitem[Krizhevsky et~al\mbox{.}(2009)]%
        {cifar}
\bibfield{author}{\bibinfo{person}{Alex Krizhevsky}, \bibinfo{person}{Geoffrey Hinton}, {et~al\mbox{.}}} \bibinfo{year}{2009}\natexlab{}.
\newblock \showarticletitle{Learning multiple layers of features from tiny images}.
\newblock  (\bibinfo{year}{2009}).
\newblock


\bibitem[Li et~al\mbox{.}(2024a)]%
        {laf}
\bibfield{author}{\bibinfo{person}{Jia Li}, \bibinfo{person}{Lijie Hu}, \bibinfo{person}{Zhixian He}, \bibinfo{person}{Jingfeng Zhang}, \bibinfo{person}{Tianhang Zheng}, {and} \bibinfo{person}{Di Wang}.} \bibinfo{year}{2024}\natexlab{a}.
\newblock \showarticletitle{Text Guided Image Editing with Automatic Concept Locating and Forgetting}.
\newblock \bibinfo{journal}{\emph{CoRR}}  \bibinfo{volume}{abs/2405.19708} (\bibinfo{year}{2024}).
\newblock
\href{https://doi.org/10.48550/ARXIV.2405.19708}{doi:\nolinkurl{10.48550/ARXIV.2405.19708}}
\showeprint[arXiv]{2405.19708}


\bibitem[Li et~al\mbox{.}(2025)]%
        {fd}
\bibfield{author}{\bibinfo{person}{Jia Li}, \bibinfo{person}{Lijie Hu}, \bibinfo{person}{Jingfeng Zhang}, \bibinfo{person}{Tianhang Zheng}, \bibinfo{person}{Hua Zhang}, {and} \bibinfo{person}{Di Wang}.} \bibinfo{year}{2025}\natexlab{}.
\newblock \showarticletitle{Fair Text-to-Image Diffusion via Fair Mapping}. In \bibinfo{booktitle}{\emph{AAAI-25, Sponsored by the Association for the Advancement of Artificial Intelligence, February 25 - March 4, 2025, Philadelphia, PA, {USA}}}, \bibfield{editor}{\bibinfo{person}{Toby Walsh}, \bibinfo{person}{Julie Shah}, {and} \bibinfo{person}{Zico Kolter}} (Eds.). \bibinfo{publisher}{{AAAI} Press}, \bibinfo{pages}{26256--26264}.
\newblock
\href{https://doi.org/10.1609/AAAI.V39I25.34823}{doi:\nolinkurl{10.1609/AAAI.V39I25.34823}}


\bibitem[Li et~al\mbox{.}(2022b)]%
        {peijian}
\bibfield{author}{\bibinfo{person}{Junyi Li}, \bibinfo{person}{Jian Pei}, {and} \bibinfo{person}{Heng Huang}.} \bibinfo{year}{2022}\natexlab{b}.
\newblock \showarticletitle{Communication-Efficient Robust Federated Learning with Noisy Labels}. In \bibinfo{booktitle}{\emph{{KDD} '22: The 28th {ACM} {SIGKDD} Conference on Knowledge Discovery and Data Mining, Washington, DC, USA, August 14 - 18, 2022}}, \bibfield{editor}{\bibinfo{person}{Aidong Zhang} {and} \bibinfo{person}{Huzefa Rangwala}} (Eds.). \bibinfo{publisher}{{ACM}}, \bibinfo{pages}{914--924}.
\newblock
\href{https://doi.org/10.1145/3534678.3539328}{doi:\nolinkurl{10.1145/3534678.3539328}}


\bibitem[Li et~al\mbox{.}(2020b)]%
        {dividemix}
\bibfield{author}{\bibinfo{person}{Junnan Li}, \bibinfo{person}{Richard Socher}, {and} \bibinfo{person}{Steven C.~H. Hoi}.} \bibinfo{year}{2020}\natexlab{b}.
\newblock \showarticletitle{DivideMix: Learning with Noisy Labels as Semi-supervised Learning}. In \bibinfo{booktitle}{\emph{8th International Conference on Learning Representations, {ICLR} 2020, Addis Ababa, Ethiopia, April 26-30, 2020}}. \bibinfo{publisher}{OpenReview.net}.
\newblock
\urldef\tempurl%
\url{https://openreview.net/forum?id=HJgExaVtwr}
\showURL{%
\tempurl}


\bibitem[Li et~al\mbox{.}(2022a)]%
        {qinbinCrossSilo}
\bibfield{author}{\bibinfo{person}{Qinbin Li}, \bibinfo{person}{Yiqun Diao}, \bibinfo{person}{Quan Chen}, {and} \bibinfo{person}{Bingsheng He}.} \bibinfo{year}{2022}\natexlab{a}.
\newblock \showarticletitle{Federated Learning on Non-IID Data Silos: An Experimental Study}. In \bibinfo{booktitle}{\emph{38th {IEEE} International Conference on Data Engineering, {ICDE} 2022, Kuala Lumpur, Malaysia, May 9-12, 2022}}. \bibinfo{publisher}{{IEEE}}, \bibinfo{pages}{965--978}.
\newblock
\href{https://doi.org/10.1109/ICDE53745.2022.00077}{doi:\nolinkurl{10.1109/ICDE53745.2022.00077}}


\bibitem[Li et~al\mbox{.}(2021a)]%
        {moon}
\bibfield{author}{\bibinfo{person}{Qinbin Li}, \bibinfo{person}{Bingsheng He}, {and} \bibinfo{person}{Dawn Song}.} \bibinfo{year}{2021}\natexlab{a}.
\newblock \showarticletitle{Model-Contrastive Federated Learning}. In \bibinfo{booktitle}{\emph{{IEEE} Conference on Computer Vision and Pattern Recognition, {CVPR} 2021, virtual, June 19-25, 2021}}. \bibinfo{publisher}{Computer Vision Foundation / {IEEE}}, \bibinfo{pages}{10713--10722}.
\newblock
\href{https://doi.org/10.1109/CVPR46437.2021.01057}{doi:\nolinkurl{10.1109/CVPR46437.2021.01057}}


\bibitem[Li et~al\mbox{.}(2021b)]%
        {median}
\bibfield{author}{\bibinfo{person}{Tian Li}, \bibinfo{person}{Shengyuan Hu}, \bibinfo{person}{Ahmad Beirami}, {and} \bibinfo{person}{Virginia Smith}.} \bibinfo{year}{2021}\natexlab{b}.
\newblock \showarticletitle{Ditto: Fair and Robust Federated Learning Through Personalization}. In \bibinfo{booktitle}{\emph{Proceedings of the 38th International Conference on Machine Learning, {ICML} 2021, 18-24 July 2021, Virtual Event}} \emph{(\bibinfo{series}{Proceedings of Machine Learning Research}, Vol.~\bibinfo{volume}{139})}, \bibfield{editor}{\bibinfo{person}{Marina Meila} {and} \bibinfo{person}{Tong Zhang}} (Eds.). \bibinfo{publisher}{{PMLR}}, \bibinfo{pages}{6357--6368}.
\newblock
\urldef\tempurl%
\url{http://proceedings.mlr.press/v139/li21h.html}
\showURL{%
\tempurl}


\bibitem[Li et~al\mbox{.}(2020a)]%
        {fedprox}
\bibfield{author}{\bibinfo{person}{Tian Li}, \bibinfo{person}{Anit~Kumar Sahu}, \bibinfo{person}{Manzil Zaheer}, \bibinfo{person}{Maziar Sanjabi}, \bibinfo{person}{Ameet Talwalkar}, {and} \bibinfo{person}{Virginia Smith}.} \bibinfo{year}{2020}\natexlab{a}.
\newblock \showarticletitle{Federated Optimization in Heterogeneous Networks}. In \bibinfo{booktitle}{\emph{Proceedings of Machine Learning and Systems 2020, MLSys 2020, Austin, TX, USA, March 2-4, 2020}}, \bibfield{editor}{\bibinfo{person}{Inderjit~S. Dhillon}, \bibinfo{person}{Dimitris~S. Papailiopoulos}, {and} \bibinfo{person}{Vivienne Sze}} (Eds.). \bibinfo{publisher}{mlsys.org}.
\newblock
\urldef\tempurl%
\url{https://proceedings.mlsys.org/book/316.pdf}
\showURL{%
\tempurl}


\bibitem[Li et~al\mbox{.}(2023)]%
        {fedtrip}
\bibfield{author}{\bibinfo{person}{Xujing Li}, \bibinfo{person}{Min Liu}, \bibinfo{person}{Sheng Sun}, \bibinfo{person}{Yuwei Wang}, \bibinfo{person}{Hui Jiang}, {and} \bibinfo{person}{Xuefeng Jiang}.} \bibinfo{year}{2023}\natexlab{}.
\newblock \showarticletitle{FedTrip: {A} Resource-Efficient Federated Learning Method with Triplet Regularization}. In \bibinfo{booktitle}{\emph{{IEEE} International Parallel and Distributed Processing Symposium, {IPDPS} 2023, St. Petersburg, FL, USA, May 15-19, 2023}}. \bibinfo{publisher}{{IEEE}}, \bibinfo{pages}{809--819}.
\newblock
\href{https://doi.org/10.1109/IPDPS54959.2023.00086}{doi:\nolinkurl{10.1109/IPDPS54959.2023.00086}}


\bibitem[Li et~al\mbox{.}(2024b)]%
        {fedcrac}
\bibfield{author}{\bibinfo{person}{Xujing Li}, \bibinfo{person}{Sheng Sun}, \bibinfo{person}{Min Liu}, \bibinfo{person}{Ju Ren}, \bibinfo{person}{Xuefeng Jiang}, {and} \bibinfo{person}{Tianliu He}.} \bibinfo{year}{2024}\natexlab{b}.
\newblock \showarticletitle{FedCRAC: Improving Federated Classification Performance on Long-Tailed Data via Classifier Representation Adjustment and Calibration}.
\newblock \bibinfo{journal}{\emph{IEEE Transactions on Mobile Computing}} (\bibinfo{year}{2024}).
\newblock


\bibitem[Liang et~al\mbox{.}(2023)]%
        {fednoisy}
\bibfield{author}{\bibinfo{person}{Siqi Liang}, \bibinfo{person}{Jintao Huang}, \bibinfo{person}{Dun Zeng}, \bibinfo{person}{Junyuan Hong}, \bibinfo{person}{Jiayu Zhou}, {and} \bibinfo{person}{Zenglin Xu}.} \bibinfo{year}{2023}\natexlab{}.
\newblock \showarticletitle{FedNoisy: Federated Noisy Label Learning Benchmark}.
\newblock \bibinfo{journal}{\emph{CoRR}}  \bibinfo{volume}{abs/2306.11650} (\bibinfo{year}{2023}).
\newblock
\href{https://doi.org/10.48550/arXiv.2306.11650}{doi:\nolinkurl{10.48550/arXiv.2306.11650}}
\showeprint[arXiv]{2306.11650}


\bibitem[Lin et~al\mbox{.}(2025)]%
        {feddshar}
\bibfield{author}{\bibinfo{person}{Ziqian Lin}, \bibinfo{person}{Xuefeng Jiang}, \bibinfo{person}{Kun Zhang}, \bibinfo{person}{Chongjun Fan}, {and} \bibinfo{person}{Yaya Liu}.} \bibinfo{year}{2025}\natexlab{}.
\newblock \showarticletitle{FedDSHAR: {A} dual-strategy federated learning approach for human activity recognition amid noise label user}.
\newblock \bibinfo{journal}{\emph{Future Gener. Comput. Syst.}}  \bibinfo{volume}{166} (\bibinfo{year}{2025}), \bibinfo{pages}{107724}.
\newblock
\href{https://doi.org/10.1016/J.FUTURE.2025.107724}{doi:\nolinkurl{10.1016/J.FUTURE.2025.107724}}


\bibitem[Liu et~al\mbox{.}(2020)]%
        {elr}
\bibfield{author}{\bibinfo{person}{Sheng Liu}, \bibinfo{person}{Jonathan Niles{-}Weed}, \bibinfo{person}{Narges Razavian}, {and} \bibinfo{person}{Carlos Fernandez{-}Granda}.} \bibinfo{year}{2020}\natexlab{}.
\newblock \showarticletitle{Early-Learning Regularization Prevents Memorization of Noisy Labels}. In \bibinfo{booktitle}{\emph{Advances in Neural Information Processing Systems 33: Annual Conference on Neural Information Processing Systems 2020, NeurIPS 2020, December 6-12, 2020, virtual}}, \bibfield{editor}{\bibinfo{person}{Hugo Larochelle}, \bibinfo{person}{Marc'Aurelio Ranzato}, \bibinfo{person}{Raia Hadsell}, \bibinfo{person}{Maria{-}Florina Balcan}, {and} \bibinfo{person}{Hsuan{-}Tien Lin}} (Eds.).
\newblock
\urldef\tempurl%
\url{https://proceedings.neurips.cc/paper/2020/hash/ea89621bee7c88b2c5be6681c8ef4906-Abstract.html}
\showURL{%
\tempurl}


\bibitem[Lu et~al\mbox{.}(2025)]%
        {fedlf}
\bibfield{author}{\bibinfo{person}{Xiuhua Lu}, \bibinfo{person}{Peng Li}, {and} \bibinfo{person}{Xuefeng Jiang}.} \bibinfo{year}{2025}\natexlab{}.
\newblock \showarticletitle{FedLF: Adaptive Logit Adjustment and Feature Optimization in Federated Long-Tailed Learning}. In \bibinfo{booktitle}{\emph{Asian Conference on Machine Learning}}. PMLR, \bibinfo{pages}{303--318}.
\newblock


\bibitem[Lu et~al\mbox{.}(2023)]%
        {fedned}
\bibfield{author}{\bibinfo{person}{Yang Lu}, \bibinfo{person}{Lin Chen}, \bibinfo{person}{Yonggang Zhang}, \bibinfo{person}{Yiliang Zhang}, \bibinfo{person}{Bo Han}, \bibinfo{person}{Yiu-ming Cheung}, {and} \bibinfo{person}{Hanzi Wang}.} \bibinfo{year}{2023}\natexlab{}.
\newblock \showarticletitle{Federated Learning with Extremely Noisy Clients via Negative Distillation}.
\newblock \bibinfo{journal}{\emph{arXiv preprint arXiv:2312.12703}} (\bibinfo{year}{2023}).
\newblock


\bibitem[McLachlan and Rathnayake(2014)]%
        {gmm}
\bibfield{author}{\bibinfo{person}{Geoffrey~J. McLachlan} {and} \bibinfo{person}{Suren~I. Rathnayake}.} \bibinfo{year}{2014}\natexlab{}.
\newblock \showarticletitle{On the number of components in a Gaussian mixture model}.
\newblock \bibinfo{journal}{\emph{WIREs Data Mining Knowl. Discov.}} \bibinfo{volume}{4}, \bibinfo{number}{5} (\bibinfo{year}{2014}), \bibinfo{pages}{341--355}.
\newblock
\href{https://doi.org/10.1002/WIDM.1135}{doi:\nolinkurl{10.1002/WIDM.1135}}


\bibitem[McMahan et~al\mbox{.}(2017)]%
        {fedavg}
\bibfield{author}{\bibinfo{person}{Brendan McMahan}, \bibinfo{person}{Eider Moore}, \bibinfo{person}{Daniel Ramage}, \bibinfo{person}{Seth Hampson}, {and} \bibinfo{person}{Blaise~Ag{\"{u}}era y Arcas}.} \bibinfo{year}{2017}\natexlab{}.
\newblock \showarticletitle{Communication-Efficient Learning of Deep Networks from Decentralized Data}. In \bibinfo{booktitle}{\emph{Proceedings of the 20th International Conference on Artificial Intelligence and Statistics, {AISTATS} 2017, 20-22 April 2017, Fort Lauderdale, FL, {USA}}} \emph{(\bibinfo{series}{Proceedings of Machine Learning Research}, Vol.~\bibinfo{volume}{54})}, \bibfield{editor}{\bibinfo{person}{Aarti Singh} {and} \bibinfo{person}{Xiaojin~(Jerry) Zhu}} (Eds.). \bibinfo{publisher}{{PMLR}}, \bibinfo{pages}{1273--1282}.
\newblock
\urldef\tempurl%
\url{http://proceedings.mlr.press/v54/mcmahan17a.html}
\showURL{%
\tempurl}


\bibitem[Menon et~al\mbox{.}(2021)]%
        {LA}
\bibfield{author}{\bibinfo{person}{Aditya~Krishna Menon}, \bibinfo{person}{Sadeep Jayasumana}, \bibinfo{person}{Ankit~Singh Rawat}, \bibinfo{person}{Himanshu Jain}, \bibinfo{person}{Andreas Veit}, {and} \bibinfo{person}{Sanjiv Kumar}.} \bibinfo{year}{2021}\natexlab{}.
\newblock \showarticletitle{Long-tail learning via logit adjustment}. In \bibinfo{booktitle}{\emph{9th International Conference on Learning Representations, {ICLR} 2021, Virtual Event, Austria, May 3-7, 2021}}. \bibinfo{publisher}{OpenReview.net}.
\newblock
\urldef\tempurl%
\url{https://openreview.net/forum?id=37nvvqkCo5}
\showURL{%
\tempurl}


\bibitem[Micikevicius et~al\mbox{.}(2018)]%
        {mixedPrecision}
\bibfield{author}{\bibinfo{person}{Paulius Micikevicius}, \bibinfo{person}{Sharan Narang}, \bibinfo{person}{Jonah Alben}, \bibinfo{person}{Gregory Diamos}, \bibinfo{person}{Erich Elsen}, \bibinfo{person}{David Garcia}, \bibinfo{person}{Boris Ginsburg}, \bibinfo{person}{Michael Houston}, \bibinfo{person}{Oleksii Kuchaiev}, \bibinfo{person}{Ganesh Venkatesh}, {et~al\mbox{.}}} \bibinfo{year}{2018}\natexlab{}.
\newblock \showarticletitle{Mixed Precision Training}. In \bibinfo{booktitle}{\emph{International Conference on Learning Representations}}.
\newblock


\bibitem[Paszke et~al\mbox{.}(2017)]%
        {pytorch}
\bibfield{author}{\bibinfo{person}{Adam Paszke}, \bibinfo{person}{Sam Gross}, \bibinfo{person}{Soumith Chintala}, \bibinfo{person}{Gregory Chanan}, \bibinfo{person}{Edward Yang}, \bibinfo{person}{Zachary DeVito}, \bibinfo{person}{Zeming Lin}, \bibinfo{person}{Alban Desmaison}, \bibinfo{person}{Luca Antiga}, {and} \bibinfo{person}{Adam Lerer}.} \bibinfo{year}{2017}\natexlab{}.
\newblock \showarticletitle{Automatic differentiation in pytorch}.
\newblock  (\bibinfo{year}{2017}).
\newblock


\bibitem[Pedregosa et~al\mbox{.}(2011)]%
        {sklearn}
\bibfield{author}{\bibinfo{person}{Fabian Pedregosa}, \bibinfo{person}{Ga{\"e}l Varoquaux}, \bibinfo{person}{Alexandre Gramfort}, \bibinfo{person}{Vincent Michel}, \bibinfo{person}{Bertrand Thirion}, \bibinfo{person}{Olivier Grisel}, \bibinfo{person}{Mathieu Blondel}, \bibinfo{person}{Peter Prettenhofer}, \bibinfo{person}{Ron Weiss}, \bibinfo{person}{Vincent Dubourg}, {et~al\mbox{.}}} \bibinfo{year}{2011}\natexlab{}.
\newblock \showarticletitle{Scikit-learn: Machine learning in Python}.
\newblock \bibinfo{journal}{\emph{the Journal of machine Learning research}}  \bibinfo{volume}{12} (\bibinfo{year}{2011}), \bibinfo{pages}{2825--2830}.
\newblock


\bibitem[Pillutla et~al\mbox{.}(2022)]%
        {rfa}
\bibfield{author}{\bibinfo{person}{Krishna Pillutla}, \bibinfo{person}{Sham~M Kakade}, {and} \bibinfo{person}{Zaid Harchaoui}.} \bibinfo{year}{2022}\natexlab{}.
\newblock \showarticletitle{Robust aggregation for federated learning}.
\newblock \bibinfo{journal}{\emph{IEEE Transactions on Signal Processing}}  \bibinfo{volume}{70} (\bibinfo{year}{2022}), \bibinfo{pages}{1142--1154}.
\newblock


\bibitem[Radford et~al\mbox{.}(2021)]%
        {clip}
\bibfield{author}{\bibinfo{person}{Alec Radford}, \bibinfo{person}{Jong~Wook Kim}, \bibinfo{person}{Chris Hallacy}, \bibinfo{person}{Aditya Ramesh}, \bibinfo{person}{Gabriel Goh}, \bibinfo{person}{Sandhini Agarwal}, \bibinfo{person}{Girish Sastry}, \bibinfo{person}{Amanda Askell}, \bibinfo{person}{Pamela Mishkin}, \bibinfo{person}{Jack Clark}, {et~al\mbox{.}}} \bibinfo{year}{2021}\natexlab{}.
\newblock \showarticletitle{Learning transferable visual models from natural language supervision}. In \bibinfo{booktitle}{\emph{International conference on machine learning}}. PMLR, \bibinfo{pages}{8748--8763}.
\newblock


\bibitem[Ramaswamy et~al\mbox{.}(2023)]%
        {dora}
\bibfield{author}{\bibinfo{person}{VikramV. Ramaswamy}, \bibinfo{person}{SingYu Lin}, \bibinfo{person}{Dora Zhao}, \bibinfo{person}{AaronB. Adcock}, \bibinfo{person}{Laurensvander Maaten}, \bibinfo{person}{Deepti Ghadiyaram}, {and} \bibinfo{person}{Olga Russakovsky}.} \bibinfo{year}{2023}\natexlab{}.
\newblock \showarticletitle{GeoDE: a Geographically Diverse Evaluation Dataset for Object Recognition}.
\newblock  (\bibinfo{date}{Jan} \bibinfo{year}{2023}).
\newblock


\bibitem[Simonyan and Zisserman(2014)]%
        {vgg}
\bibfield{author}{\bibinfo{person}{Karen Simonyan} {and} \bibinfo{person}{Andrew Zisserman}.} \bibinfo{year}{2014}\natexlab{}.
\newblock \showarticletitle{Very deep convolutional networks for large-scale image recognition}.
\newblock \bibinfo{journal}{\emph{arXiv preprint arXiv:1409.1556}} (\bibinfo{year}{2014}).
\newblock


\bibitem[Song et~al\mbox{.}(2019)]%
        {selfie}
\bibfield{author}{\bibinfo{person}{Hwanjun Song}, \bibinfo{person}{Minseok Kim}, {and} \bibinfo{person}{Jae{-}Gil Lee}.} \bibinfo{year}{2019}\natexlab{}.
\newblock \showarticletitle{{SELFIE:} Refurbishing Unclean Samples for Robust Deep Learning}. In \bibinfo{booktitle}{\emph{Proceedings of the 36th International Conference on Machine Learning, {ICML} 2019, 9-15 June 2019, Long Beach, California, {USA}}} \emph{(\bibinfo{series}{Proceedings of Machine Learning Research}, Vol.~\bibinfo{volume}{97})}, \bibfield{editor}{\bibinfo{person}{Kamalika Chaudhuri} {and} \bibinfo{person}{Ruslan Salakhutdinov}} (Eds.). \bibinfo{publisher}{{PMLR}}, \bibinfo{pages}{5907--5915}.
\newblock
\urldef\tempurl%
\url{http://proceedings.mlr.press/v97/song19b.html}
\showURL{%
\tempurl}


\bibitem[Tahmasebian et~al\mbox{.}(2022)]%
        {robustfed}
\bibfield{author}{\bibinfo{person}{Farnaz Tahmasebian}, \bibinfo{person}{Jian Lou}, {and} \bibinfo{person}{Li Xiong}.} \bibinfo{year}{2022}\natexlab{}.
\newblock \showarticletitle{RobustFed: {A} Truth Inference Approach for Robust Federated Learning}. In \bibinfo{booktitle}{\emph{Proceedings of the 31st {ACM} International Conference on Information {\&} Knowledge Management, Atlanta, GA, USA, October 17-21, 2022}}, \bibfield{editor}{\bibinfo{person}{Mohammad~Al Hasan} {and} \bibinfo{person}{Li~Xiong}} (Eds.). \bibinfo{publisher}{{ACM}}, \bibinfo{pages}{1868--1877}.
\newblock
\href{https://doi.org/10.1145/3511808.3557439}{doi:\nolinkurl{10.1145/3511808.3557439}}


\bibitem[Tanaka et~al\mbox{.}(2018)]%
        {jointopt}
\bibfield{author}{\bibinfo{person}{Daiki Tanaka}, \bibinfo{person}{Daiki Ikami}, \bibinfo{person}{Toshihiko Yamasaki}, {and} \bibinfo{person}{Kiyoharu Aizawa}.} \bibinfo{year}{2018}\natexlab{}.
\newblock \showarticletitle{Joint Optimization Framework for Learning With Noisy Labels}. In \bibinfo{booktitle}{\emph{2018 {IEEE} Conference on Computer Vision and Pattern Recognition, {CVPR} 2018, Salt Lake City, UT, USA, June 18-22, 2018}}. \bibinfo{publisher}{Computer Vision Foundation / {IEEE} Computer Society}, \bibinfo{pages}{5552--5560}.
\newblock
\href{https://doi.org/10.1109/CVPR.2018.00582}{doi:\nolinkurl{10.1109/CVPR.2018.00582}}


\bibitem[Tsouvalas et~al\mbox{.}(2023)]%
        {chaos}
\bibfield{author}{\bibinfo{person}{Vasileios Tsouvalas}, \bibinfo{person}{Aaqib Saeed}, \bibinfo{person}{Tanir Ozcelebi}, {and} \bibinfo{person}{Nirvana Meratnia}.} \bibinfo{year}{2023}\natexlab{}.
\newblock \showarticletitle{Labeling Chaos to Learning Harmony: Federated Learning with Noisy Labels}.
\newblock \bibinfo{journal}{\emph{ACM Transactions on Intelligent Systems and Technology}} (\bibinfo{year}{2023}).
\newblock


\bibitem[Wang et~al\mbox{.}(2023)]%
        {logitsfusion}
\bibfield{author}{\bibinfo{person}{Yuwei Wang}, \bibinfo{person}{Runhan Li}, \bibinfo{person}{Hao Tan}, \bibinfo{person}{Xuefeng Jiang}, \bibinfo{person}{Sheng Sun}, \bibinfo{person}{Min Liu}, \bibinfo{person}{Bo Gao}, {and} \bibinfo{person}{Zhiyuan Wu}.} \bibinfo{year}{2023}\natexlab{}.
\newblock \showarticletitle{Federated skewed label learning with logits fusion}.
\newblock \bibinfo{journal}{\emph{arXiv preprint arXiv:2311.08202}} (\bibinfo{year}{2023}).
\newblock


\bibitem[Wang et~al\mbox{.}(2019)]%
        {symmetricce}
\bibfield{author}{\bibinfo{person}{Yisen Wang}, \bibinfo{person}{Xingjun Ma}, \bibinfo{person}{Zaiyi Chen}, \bibinfo{person}{Yuan Luo}, \bibinfo{person}{Jinfeng Yi}, {and} \bibinfo{person}{James Bailey}.} \bibinfo{year}{2019}\natexlab{}.
\newblock \showarticletitle{Symmetric Cross Entropy for Robust Learning With Noisy Labels}. In \bibinfo{booktitle}{\emph{2019 {IEEE/CVF} International Conference on Computer Vision, {ICCV} 2019, Seoul, Korea (South), October 27 - November 2, 2019}}. \bibinfo{publisher}{{IEEE}}, \bibinfo{pages}{322--330}.
\newblock
\href{https://doi.org/10.1109/ICCV.2019.00041}{doi:\nolinkurl{10.1109/ICCV.2019.00041}}


\bibitem[Wei et~al\mbox{.}(2022)]%
        {cifarn}
\bibfield{author}{\bibinfo{person}{Jiaheng Wei}, \bibinfo{person}{Zhaowei Zhu}, \bibinfo{person}{Hao Cheng}, \bibinfo{person}{Tongliang Liu}, \bibinfo{person}{Gang Niu}, {and} \bibinfo{person}{Yang Liu}.} \bibinfo{year}{2022}\natexlab{}.
\newblock \showarticletitle{Learning with Noisy Labels Revisited: {A} Study Using Real-World Human Annotations}. In \bibinfo{booktitle}{\emph{The Tenth International Conference on Learning Representations, {ICLR} 2022, Virtual Event, April 25-29, 2022}}. \bibinfo{publisher}{OpenReview.net}.
\newblock
\urldef\tempurl%
\url{https://openreview.net/forum?id=TBWA6PLJZQm}
\showURL{%
\tempurl}


\bibitem[Wen et~al\mbox{.}(2024)]%
        {wentian}
\bibfield{author}{\bibinfo{person}{Tian Wen}, \bibinfo{person}{Hanqing Zhang}, \bibinfo{person}{Han Zhang}, \bibinfo{person}{Huixin Wu}, \bibinfo{person}{Danxin Wang}, \bibinfo{person}{Xiuwen Liu}, \bibinfo{person}{Weishan Zhang}, \bibinfo{person}{Yuwei Wang}, {and} \bibinfo{person}{Shaohua Cao}.} \bibinfo{year}{2024}\natexlab{}.
\newblock \showarticletitle{RTIFed: A Reputation based Triple-step Incentive mechanism for energy-aware Federated learning over battery-constricted devices}.
\newblock \bibinfo{journal}{\emph{Computer Networks}}  \bibinfo{volume}{241} (\bibinfo{year}{2024}), \bibinfo{pages}{110192}.
\newblock


\bibitem[Wu et~al\mbox{.}(2024)]%
        {fedaaai}
\bibfield{author}{\bibinfo{person}{Nannan Wu}, \bibinfo{person}{Zhaobin Sun}, \bibinfo{person}{Zengqiang Yan}, {and} \bibinfo{person}{Li Yu}.} \bibinfo{year}{2024}\natexlab{}.
\newblock \showarticletitle{FedA3I: annotation quality-aware aggregation for federated medical image segmentation against heterogeneous annotation noise}. In \bibinfo{booktitle}{\emph{Proceedings of the AAAI Conference on Artificial Intelligence}}, Vol.~\bibinfo{volume}{38}. \bibinfo{pages}{15943--15951}.
\newblock


\bibitem[Wu et~al\mbox{.}(2023)]%
        {noro}
\bibfield{author}{\bibinfo{person}{Nannan Wu}, \bibinfo{person}{Li Yu}, \bibinfo{person}{Xuefeng Jiang}, \bibinfo{person}{Kwang-Ting Cheng}, {and} \bibinfo{person}{Zengqiang Yan}.} \bibinfo{year}{2023}\natexlab{}.
\newblock \showarticletitle{Fednoro: Towards noise-robust federated learning by addressing class imbalance and label noise heterogeneity}.
\newblock \bibinfo{journal}{\emph{arXiv preprint arXiv:2305.05230}} (\bibinfo{year}{2023}).
\newblock


\bibitem[Xiang et~al\mbox{.}(2024)]%
        {fedia}
\bibfield{author}{\bibinfo{person}{Yangyang Xiang}, \bibinfo{person}{Nannan Wu}, \bibinfo{person}{Li Yu}, \bibinfo{person}{Xin Yang}, \bibinfo{person}{Kwang-Ting Cheng}, {and} \bibinfo{person}{Zengqiang Yan}.} \bibinfo{year}{2024}\natexlab{}.
\newblock \showarticletitle{FedIA: Federated Medical Image Segmentation with Heterogeneous Annotation Completeness}. In \bibinfo{booktitle}{\emph{International Conference on Medical Image Computing and Computer-Assisted Intervention}}. Springer, \bibinfo{pages}{373--382}.
\newblock


\bibitem[Xiao et~al\mbox{.}(2015)]%
        {clothing1m}
\bibfield{author}{\bibinfo{person}{Tong Xiao}, \bibinfo{person}{Tian Xia}, \bibinfo{person}{Yi Yang}, \bibinfo{person}{Chang Huang}, {and} \bibinfo{person}{Xiaogang Wang}.} \bibinfo{year}{2015}\natexlab{}.
\newblock \showarticletitle{Learning from massive noisy labeled data for image classification}. In \bibinfo{booktitle}{\emph{{IEEE} Conference on Computer Vision and Pattern Recognition, {CVPR} 2015, Boston, MA, USA, June 7-12, 2015}}. \bibinfo{publisher}{{IEEE} Computer Society}, \bibinfo{pages}{2691--2699}.
\newblock
\href{https://doi.org/10.1109/CVPR.2015.7298885}{doi:\nolinkurl{10.1109/CVPR.2015.7298885}}


\bibitem[Xu et~al\mbox{.}(2022)]%
        {fedcorr}
\bibfield{author}{\bibinfo{person}{Jingyi Xu}, \bibinfo{person}{Zihan Chen}, \bibinfo{person}{Tony Q.~S. Quek}, {and} \bibinfo{person}{Kai Fong~Ernest Chong}.} \bibinfo{year}{2022}\natexlab{}.
\newblock \showarticletitle{FedCorr: Multi-Stage Federated Learning for Label Noise Correction}. In \bibinfo{booktitle}{\emph{{IEEE/CVF} Conference on Computer Vision and Pattern Recognition, {CVPR} 2022, New Orleans, LA, USA, June 18-24, 2022}}. \bibinfo{publisher}{{IEEE}}, \bibinfo{pages}{10174--10183}.
\newblock
\href{https://doi.org/10.1109/CVPR52688.2022.00994}{doi:\nolinkurl{10.1109/CVPR52688.2022.00994}}


\bibitem[Xue et~al\mbox{.}(2023)]%
        {fedbiad}
\bibfield{author}{\bibinfo{person}{Jingjing Xue}, \bibinfo{person}{Min Liu}, \bibinfo{person}{Sheng Sun}, \bibinfo{person}{Yuwei Wang}, \bibinfo{person}{Hui Jiang}, {and} \bibinfo{person}{Xuefeng Jiang}.} \bibinfo{year}{2023}\natexlab{}.
\newblock \showarticletitle{FedBIAD: Communication-Efficient and Accuracy-Guaranteed Federated Learning with Bayesian Inference-Based Adaptive Dropout}. In \bibinfo{booktitle}{\emph{{IEEE} International Parallel and Distributed Processing Symposium, {IPDPS} 2023, St. Petersburg, FL, USA, May 15-19, 2023}}. \bibinfo{publisher}{{IEEE}}, \bibinfo{pages}{489--500}.
\newblock
\href{https://doi.org/10.1109/IPDPS54959.2023.00056}{doi:\nolinkurl{10.1109/IPDPS54959.2023.00056}}


\bibitem[Xue et~al\mbox{.}(2025)]%
        {jingjing}
\bibfield{author}{\bibinfo{person}{Jingjing Xue}, \bibinfo{person}{Sheng Sun}, \bibinfo{person}{Min Liu}, \bibinfo{person}{Qi Li}, {and} \bibinfo{person}{Ke Xu}.} \bibinfo{year}{2025}\natexlab{}.
\newblock \showarticletitle{Enhancing Federated Learning Robustness using Locally Benignity-Assessable Bayesian Dropout}.
\newblock \bibinfo{journal}{\emph{IEEE Transactions on Information Forensics and Security}} (\bibinfo{year}{2025}).
\newblock


\bibitem[Yan et~al\mbox{.}(2024a)]%
        {fedeye}
\bibfield{author}{\bibinfo{person}{Bingjie Yan}, \bibinfo{person}{Danmin Cao}, \bibinfo{person}{Xinlong Jiang}, \bibinfo{person}{Yiqiang Chen}, \bibinfo{person}{Weiwei Dai}, \bibinfo{person}{Fan Dong}, \bibinfo{person}{Wuliang Huang}, \bibinfo{person}{Teng Zhang}, \bibinfo{person}{Chenlong Gao}, \bibinfo{person}{Qian Chen}, {et~al\mbox{.}}} \bibinfo{year}{2024}\natexlab{a}.
\newblock \showarticletitle{FedEYE: A scalable and flexible end-to-end federated learning platform for ophthalmology}.
\newblock \bibinfo{journal}{\emph{Patterns}} \bibinfo{volume}{5}, \bibinfo{number}{2} (\bibinfo{year}{2024}).
\newblock


\bibitem[Yan et~al\mbox{.}(2024b)]%
        {bingjie}
\bibfield{author}{\bibinfo{person}{Bingjie Yan}, \bibinfo{person}{Qian Chen}, \bibinfo{person}{Yiqiang Chen}, \bibinfo{person}{Xinlong Jiang}, \bibinfo{person}{Wuliang Huang}, \bibinfo{person}{Bingyu Wang}, \bibinfo{person}{Zhirui Wang}, \bibinfo{person}{Chenlong Gao}, {and} \bibinfo{person}{Teng Zhang}.} \bibinfo{year}{2024}\natexlab{b}.
\newblock \showarticletitle{Buffalo: Biomedical Vision-Language Understanding with Cross-Modal Prototype and Federated Foundation Model Collaboration}. In \bibinfo{booktitle}{\emph{Proceedings of the 33rd ACM International Conference on Information and Knowledge Management}}. \bibinfo{pages}{2775--2785}.
\newblock


\bibitem[Yang et~al\mbox{.}(2021)]%
        {smartphone}
\bibfield{author}{\bibinfo{person}{Chengxu Yang}, \bibinfo{person}{Qipeng Wang}, \bibinfo{person}{Mengwei Xu}, \bibinfo{person}{Zhenpeng Chen}, \bibinfo{person}{Kaigui Bian}, \bibinfo{person}{Yunxin Liu}, {and} \bibinfo{person}{Xuanzhe Liu}.} \bibinfo{year}{2021}\natexlab{}.
\newblock \showarticletitle{Characterizing Impacts of Heterogeneity in Federated Learning upon Large-Scale Smartphone Data}. In \bibinfo{booktitle}{\emph{{WWW} '21: The Web Conference 2021, Virtual Event / Ljubljana, Slovenia, April 19-23, 2021}}, \bibfield{editor}{\bibinfo{person}{Jure Leskovec}, \bibinfo{person}{Marko Grobelnik}, \bibinfo{person}{Marc Najork}, \bibinfo{person}{Jie Tang}, {and} \bibinfo{person}{Leila Zia}} (Eds.). \bibinfo{publisher}{{ACM} / {IW3C2}}, \bibinfo{pages}{935--946}.
\newblock
\href{https://doi.org/10.1145/3442381.3449851}{doi:\nolinkurl{10.1145/3442381.3449851}}


\bibitem[Yang et~al\mbox{.}(2022)]%
        {robustfl}
\bibfield{author}{\bibinfo{person}{Seunghan Yang}, \bibinfo{person}{Hyoungseob Park}, \bibinfo{person}{Junyoung Byun}, {and} \bibinfo{person}{Changick Kim}.} \bibinfo{year}{2022}\natexlab{}.
\newblock \showarticletitle{Robust Federated Learning With Noisy Labels}.
\newblock \bibinfo{journal}{\emph{{IEEE} Intell. Syst.}} \bibinfo{volume}{37}, \bibinfo{number}{2} (\bibinfo{year}{2022}), \bibinfo{pages}{35--43}.
\newblock
\href{https://doi.org/10.1109/MIS.2022.3151466}{doi:\nolinkurl{10.1109/MIS.2022.3151466}}


\bibitem[Ye et~al\mbox{.}(2023)]%
        {yerui}
\bibfield{author}{\bibinfo{person}{Rui Ye}, \bibinfo{person}{Yaxin Du}, \bibinfo{person}{Zhenyang Ni}, \bibinfo{person}{Siheng Chen}, {and} \bibinfo{person}{Yanfeng Wang}.} \bibinfo{year}{2023}\natexlab{}.
\newblock \showarticletitle{Fake it till make it: Federated learning with consensus-oriented generation}.
\newblock \bibinfo{journal}{\emph{arXiv preprint arXiv:2312.05966}} (\bibinfo{year}{2023}).
\newblock


\bibitem[Yi and Wu(2019)]%
        {pencil}
\bibfield{author}{\bibinfo{person}{Kun Yi} {and} \bibinfo{person}{Jianxin Wu}.} \bibinfo{year}{2019}\natexlab{}.
\newblock \showarticletitle{Probabilistic end-to-end noise correction for learning with noisy labels}. In \bibinfo{booktitle}{\emph{Proceedings of the IEEE/CVF conference on computer vision and pattern recognition}}. \bibinfo{pages}{7017--7025}.
\newblock


\bibitem[Yin et~al\mbox{.}(2018)]%
        {trimmedmean}
\bibfield{author}{\bibinfo{person}{Dong Yin}, \bibinfo{person}{Yudong Chen}, \bibinfo{person}{Kannan Ramchandran}, {and} \bibinfo{person}{Peter~L. Bartlett}.} \bibinfo{year}{2018}\natexlab{}.
\newblock \showarticletitle{Byzantine-Robust Distributed Learning: Towards Optimal Statistical Rates}. In \bibinfo{booktitle}{\emph{Proceedings of the 35th International Conference on Machine Learning, {ICML} 2018, Stockholmsm{\"{a}}ssan, Stockholm, Sweden, July 10-15, 2018}} \emph{(\bibinfo{series}{Proceedings of Machine Learning Research}, Vol.~\bibinfo{volume}{80})}, \bibfield{editor}{\bibinfo{person}{Jennifer~G. Dy} {and} \bibinfo{person}{Andreas Krause}} (Eds.). \bibinfo{publisher}{{PMLR}}, \bibinfo{pages}{5636--5645}.
\newblock
\urldef\tempurl%
\url{http://proceedings.mlr.press/v80/yin18a.html}
\showURL{%
\tempurl}


\bibitem[Yu et~al\mbox{.}(2019)]%
        {coteaching+}
\bibfield{author}{\bibinfo{person}{Xingrui Yu}, \bibinfo{person}{Bo Han}, \bibinfo{person}{Jiangchao Yao}, \bibinfo{person}{Gang Niu}, \bibinfo{person}{Ivor~W. Tsang}, {and} \bibinfo{person}{Masashi Sugiyama}.} \bibinfo{year}{2019}\natexlab{}.
\newblock \showarticletitle{How does Disagreement Help Generalization against Label Corruption?}. In \bibinfo{booktitle}{\emph{Proceedings of the 36th International Conference on Machine Learning, {ICML} 2019, 9-15 June 2019, Long Beach, California, {USA}}} \emph{(\bibinfo{series}{Proceedings of Machine Learning Research}, Vol.~\bibinfo{volume}{97})}, \bibfield{editor}{\bibinfo{person}{Kamalika Chaudhuri} {and} \bibinfo{person}{Ruslan Salakhutdinov}} (Eds.). \bibinfo{publisher}{{PMLR}}, \bibinfo{pages}{7164--7173}.
\newblock
\urldef\tempurl%
\url{http://proceedings.mlr.press/v97/yu19b.html}
\showURL{%
\tempurl}


\bibitem[Zhang et~al\mbox{.}(2021)]%
        {incentives}
\bibfield{author}{\bibinfo{person}{Jingwen Zhang}, \bibinfo{person}{Yuezhou Wu}, {and} \bibinfo{person}{Rong Pan}.} \bibinfo{year}{2021}\natexlab{}.
\newblock \showarticletitle{Incentive Mechanism for Horizontal Federated Learning Based on Reputation and Reverse Auction}. In \bibinfo{booktitle}{\emph{{WWW} '21: The Web Conference 2021, Virtual Event / Ljubljana, Slovenia, April 19-23, 2021}}, \bibfield{editor}{\bibinfo{person}{Jure Leskovec}, \bibinfo{person}{Marko Grobelnik}, \bibinfo{person}{Marc Najork}, \bibinfo{person}{Jie Tang}, {and} \bibinfo{person}{Leila Zia}} (Eds.). \bibinfo{publisher}{{ACM} / {IW3C2}}, \bibinfo{pages}{947--956}.
\newblock
\href{https://doi.org/10.1145/3442381.3449888}{doi:\nolinkurl{10.1145/3442381.3449888}}


\bibitem[Zhang et~al\mbox{.}(2015)]%
        {agnews}
\bibfield{author}{\bibinfo{person}{Xiang Zhang}, \bibinfo{person}{Junbo~Jake Zhao}, {and} \bibinfo{person}{Yann LeCun}.} \bibinfo{year}{2015}\natexlab{}.
\newblock \showarticletitle{Character-level Convolutional Networks for Text Classification}. In \bibinfo{booktitle}{\emph{Advances in Neural Information Processing Systems 28: Annual Conference on Neural Information Processing Systems 2015, December 7-12, 2015, Montreal, Quebec, Canada}}, \bibfield{editor}{\bibinfo{person}{Corinna Cortes}, \bibinfo{person}{Neil~D. Lawrence}, \bibinfo{person}{Daniel~D. Lee}, \bibinfo{person}{Masashi Sugiyama}, {and} \bibinfo{person}{Roman Garnett}} (Eds.). \bibinfo{pages}{649--657}.
\newblock
\urldef\tempurl%
\url{https://proceedings.neurips.cc/paper/2015/hash/250cf8b51c773f3f8dc8b4be867a9a02-Abstract.html}
\showURL{%
\tempurl}


\bibitem[Zhang et~al\mbox{.}(2024)]%
        {zhiqin}
\bibfield{author}{\bibinfo{person}{Yonggang Zhang}, \bibinfo{person}{Zhiqin Yang}, \bibinfo{person}{Xinmei Tian}, \bibinfo{person}{Nannan Wang}, \bibinfo{person}{Tongliang Liu}, {and} \bibinfo{person}{Bo Han}.} \bibinfo{year}{2024}\natexlab{}.
\newblock \showarticletitle{Robust Training of Federated Models with Extremely Label Deficiency}. In \bibinfo{booktitle}{\emph{The Twelfth International Conference on Learning Representations}}.
\newblock
\urldef\tempurl%
\url{https://openreview.net/forum?id=qxLVaYbsSI}
\showURL{%
\tempurl}


\bibitem[Zhang and Sabuncu(2018)]%
        {gce}
\bibfield{author}{\bibinfo{person}{Zhilu Zhang} {and} \bibinfo{person}{Mert~R. Sabuncu}.} \bibinfo{year}{2018}\natexlab{}.
\newblock \showarticletitle{Generalized Cross Entropy Loss for Training Deep Neural Networks with Noisy Labels}. In \bibinfo{booktitle}{\emph{Advances in Neural Information Processing Systems 31: Annual Conference on Neural Information Processing Systems 2018, NeurIPS 2018, December 3-8, 2018, Montr{\'{e}}al, Canada}}, \bibfield{editor}{\bibinfo{person}{Samy Bengio}, \bibinfo{person}{Hanna~M. Wallach}, \bibinfo{person}{Hugo Larochelle}, \bibinfo{person}{Kristen Grauman}, \bibinfo{person}{Nicol{\`{o}} Cesa{-}Bianchi}, {and} \bibinfo{person}{Roman Garnett}} (Eds.). \bibinfo{pages}{8792--8802}.
\newblock
\urldef\tempurl%
\url{https://proceedings.neurips.cc/paper/2018/hash/f2925f97bc13ad2852a7a551802feea0-Abstract.html}
\showURL{%
\tempurl}


\end{thebibliography}

\end{document}